\documentclass[10pt, a4paper, twocolumn]{article} 

\usepackage[utf8]{inputenc}
\usepackage[T1]{fontenc}
\usepackage[english]{babel}

\usepackage{tgheros} 

\usepackage[varqu]{zi4} 
\usepackage{microtype}  

\usepackage{titlesec}
\titlespacing*{\section}{0pt}{1.5ex plus 1ex minus .2ex}{0.1em} 
\titlespacing*{\subsection}{0pt}{1.5ex plus 1ex minus .2ex}{0.1em}
\titlespacing*{\subsubsection}{0pt}{1.5ex plus 1ex minus .2ex}{0.1em}

\usepackage[table]{xcolor} 
\usepackage{booktabs} 
\usepackage{float}           
\usepackage{graphicx}
\usepackage{subfig}
\usepackage[labelfont={bf,small}, textfont={small}]{caption} 

\definecolor{acqua_celeste}{RGB}{0, 160, 175}

\usepackage{amsmath}
\usepackage{amssymb}
\usepackage{mathtools}
\usepackage{amsthm}
\usepackage{bbold}     
\usepackage{pifont}     
\usepackage{multirow}
\usepackage{arydshln}   
\usepackage{array}
\usepackage{makecell}   
\usepackage{wrapfig}    
\usepackage{tabularx}   

\definecolor{pgreen}{rgb}{0.13, 0.55, 0.13}
\definecolor{pred}{rgb}{0.8, 0.13, 0.13}
\definecolor{diversity}{HTML}{D4E1F5}
\definecolor{recognizability}{HTML}{FFE6CC}

\newcommand{\xmark}{\ding{55}}

\newcolumntype{Y}{>{\centering\arraybackslash}X}

\theoremstyle{plain}

\theoremstyle{definition}

\theoremstyle{remark}

\usepackage[left=1.5cm, right=1.5cm, top=1.5cm, bottom=2cm]{geometry}
\usepackage[numbers, sort&compress]{natbib}
\usepackage{hyperref}
\usepackage[capitalize,noabbrev]{cleveref} 

\hypersetup{
    colorlinks=true,        
    linkcolor=black,        
    urlcolor=acqua_celeste, 
    citecolor=acqua_celeste 
}

\newcommand{\citea}[1]{{\hypersetup{citecolor=black}\citeauthor{#1}}~\cite{#1}}

\usepackage{fancyhdr}
\usepackage{authblk}

\title{\huge\bfseries\vspace{-1em} PolyGen: Fully Synthetic Vision-Language Training via Multi-Generator Ensembles}

\author[1]{Leonardo Brusini}
\author[1]{Cristian Sbrolli\textsuperscript{*}}
\author[1]{Eugenio Lomurno}
\author[2]{\authorcr Toshihiko Yamasaki} 
\author[1]{Matteo Matteucci}

\affil[1]{Politecnico di Milano, AIRLab, Italy}
\affil[2]{University of Tokyo, Computer Vision and Media Lab, Japan}

\date{}

\begin{document}

\twocolumn[
  \begin{@twocolumnfalse}
    \maketitle
    \vspace{-3em} 

    \begin{abstract}
        \setlength{\parindent}{0pt}
        \setlength{\parskip}{4pt}
        \itshape
        \noindent Synthetic data offers a scalable solution for vision-language pre-training, yet current state-of-the-art methods typically rely on scaling up a single generative backbone, which introduces generator-specific spectral biases and limits feature diversity.
        In this work, we introduce \textbf{PolyGen}, a framework that redefines synthetic data construction by prioritizing \textit{manifold coverage} and \textit{compositional rigor} over simple dataset size. PolyGen employs a Polylithic approach to train on the intersection of architecturally distinct generators, effectively marginalizing out model-specific artifacts. Additionally, we introduce a Programmatic Hard Negative curriculum that enforces fine-grained syntactic understanding. By structurally reallocating the same data budget from unique captions to multi-source variations, PolyGen achieves a more robust feature space, outperforming the leading single-source baseline (SynthCLIP) by +19.0\% on aggregate multi-task benchmarks and on the SugarCrepe++ compositionality benchmark (+9.1\%). These results demonstrate that structural diversity is a more data-efficient scaling law than simply increasing the volume of single-source samples.
    \end{abstract}

\vspace{0.5em}

    \vspace{1em}
    \hrule height 1pt 
    \vspace{2em} 
  \end{@twocolumnfalse}
]

{
  \renewcommand{\thefootnote}{\fnsymbol{footnote}} 
  \footnotetext[1]{Corresponding author: \texttt{cristian.sbrolli@polimi.it}}
}


\section{Introduction}\label{sec:introduction}
One of the major breakthroughs in artificial intelligence has been the emergence of Vision-Language Models (VLMs) capable of aligning visual and textual representations in a shared embedding space. Models such as CLIP \citep{radford2021learning} and ALIGN \citep{jia2021scaling} have unlocked remarkable zero-shot capabilities, effectively decoupling visual recognition from the constraints of fixed label sets. However, the efficacy of these models rests upon a foundation that is becoming increasingly precarious: massive, web-scale datasets comprising billions of image-text pairs.

This reliance on web-scraped data presents two existential challenges. First, recent work has documented concerning trends suggesting that the supply of high-quality, human-generated data may become a bottleneck for future scaling \citep{villalobos2022will}. Second, ``real-world'' datasets are inherently noisy, rife with misalignment, societal biases, and privacy or copyright issues that require expensive curation. A compelling alternative is the transition to fully synthetic data \citep{hammoud2024synthclip, tian2024learning, lomurno2024stable, lomurno2025synthetic}. Given the fidelity of modern text-to-image (T2I) diffusion models, it is theoretically possible to generate infinitely scalable, safe, and controllable datasets on demand \citep{fan2024scaling}.

Despite this promise, a persistent ``Synthetic Gap'' remains: models trained exclusively on synthetic data consistently underperform their real-data counterparts, particularly in generalization tasks. Recent work by \citet{fan2024scaling} has identified a fundamental cause: a trade-off between \textit{diversity} and \textit{recognizability} that is inherent to single-generator training. Moreover, empirical evidence suggests that even ensembles of identical models trained on the same data can outperform single-model approaches, revealing that individual generators are confined to local optima in the manifold space \citep{lampis2023bridging, lomurno2025federated}. 

The manifestation of this trade-off varies by architecture. Older diffusion models (e.g., Stable Diffusion 1.5) produce high semantic variance but suffer from perceptual artifacts and frequency-domain signatures, while modern models (e.g., SDXL-Turbo) achieve higher photorealism at the expense of reduced output diversity. When trained on images from a single generator, VLMs inevitably learn to rely on these low-level, generator-specific cues, such as texture patterns or spectral fingerprints \citep{corvi2023detection}, rather than the abstract semantic features necessary for open-world generalization.

To address these challenges, we introduce \textbf{PolyGen} (Polylithic Generation Framework), an end-to-end pipeline for training vision-language models on fully synthetic data. Our approach is grounded in the \textit{Generator-Invariance Hypothesis}: by training on images sampled from multiple architecturally distinct generators, we force the model to learn representations that are invariant to generator-specific artifacts, effectively marginalizing out spurious correlations and isolating the semantic signal as the learning objective.

We adapt our approach to also generate multiple hard negative samples, addressing a separate limitation of contrastive learning: standard training objectives often reduce to ``bag-of-words'' matching, where models identify objects through simple co-occurrence statistics rather than fine grained compositional understanding \citep{yuksekgonul2022and}. 

Finally, PolyGen operationalizes the Generator-Invariance hypothesis through three complementary mechanisms:
\begin{enumerate}
    \item \textbf{Multi-Positive Ensemble:} We break the single-generator trade-off by generating $n^+$ diverse visualizations for each caption using a heterogeneous pool (SD 1.5, SD 2, SDXL-Turbo, SANA). By treating these architecturally distinct outputs as positive pairs through multi-positive objectives, we force the model to learn representations invariant to specific spectral artifacts but sensitive to the shared semantic content.
    \item \textbf{Programmatic Hard Negatives:} We leverage LLMs to construct structured and controlled semantic perturbations (counterfactuals) for the base captions. Training on these hard negatives forces the model to abandon ``bag-of-words'' shortcuts in favor of fine-grained compositional understanding, distinguishing between different semantic and compositional bindings.
    \item \textbf{Curriculum Scheduler:} We gradually increase the proportion of hard negatives during training, allowing the model to first establish coarse semantic structure before refining discriminative boundaries.
\end{enumerate}

Empirically, PolyGen demonstrates that balancing synthetic data diversity and quality is crucial. By engineering the training distribution rather than merely scaling it, we achieve a \textbf{+19.0\%} improvement in Delta Multi-Task Learning over strong single-source baselines, narrowing the gap to real-data performance and suggesting a viable path toward sustainable large-scale vision-language training.

\section{Related Work}\label{sec:related_work}
\textbf{Hybrid Synthetic Strategies.}
A dominant line of research mitigates the data scarcity problem by leveraging real-world captions to generate synthetic images. StableRep \citep{tian2023stablerep} demonstrated that multi-positive contrastive learning, treating multiple synthetic images generated from the same caption as positive pairs, can outperform models trained on real data. LaCLIP \citep{fan2023improving} and VeCLIP \citep{lai2024veclip} use instead Large Language Models (LLMs) to rewrite captions, improving alignment and diversity while maintaining the original image-text grounding. However, these ``hybrid'' approaches remain tethered to the constraints of web-scraped data, inheriting its linguistic/visual biases, privacy risks, and finite scale.

\textbf{Fully Synthetic Pipelines.}
Recent works have attempted to close the loop by generating both captions and images \citep{tian2024learning}. \citet{fan2024scaling} proposed scaling laws for synthetic data, showing that synthetic pre-training can rival real-world baselines. A landmark study, SynthCLIP \citep{hammoud2024synthclip}, introduced an end-to-end pipeline using LLM-generated captions and Stable Diffusion images, training a CLIP model from scratch without any human data. Recent work has further demonstrated that synthetic data can achieve parity with real datasets in image classification tasks while providing privacy guarantees \citep{resmini2025your}, though this result remains limited to single-task supervised learning. While these approaches are pioneering, they suffer from \textit{architectural homogeneity}: they typically rely on a single generative model, often relying on dated architectures (e.g., Stable Diffusion v1.5), causing the downstream model to overfit to generator-specific artifacts \citep{corvi2023detection}. We posit that this lack of generator diversity is a primary contributor to the persistent ``Synthetic Gap'' in vision-language tasks, as models learn to exploit low-level texture patterns rather than abstract semantic features.

\textbf{Hard Negative Mining.}
A well-known limitation of contrastive vision-language models is their tendency to rely on lexical co-occurrence patterns rather than true compositional reasoning \citep{yuksekgonul2022and}. Benchmarks like ARO and SugarCrepe \citep{hsieh2023sugarcrepe} systematically evaluate this failure mode through tasks requiring attribute binding, spatial reasoning, and object-relation understanding, revealing significant performance gaps even in state-of-the-art models. To enhance discriminative capability, Hard Negative Mining has evolved from importance weighting \citep{robinson2020contrastive,radenovic2023filtering}, to hard negative generation with rule-based swapping \citep{yuksekgonul2022and} and LLM-driven rewriting \citep{patel2024tripletclip}. However, existing methods either depend on external parsers or apply unconstrained LLM generation to unstructured text, risking semantic drift. While intrinsic data mining has been proposed \citep{wang2025getting}, it remains computationally expensive and bound by the static data distribution. We propose a \textit{constructive} alternative: by synthesizing structurally grounded base captions first, we can programmatically force the LLM to perturb specific semantic axes, ensuring safer, high-fidelity counterfactuals.

\begin{figure*}[t!]
    \centering
    \includegraphics[width=\textwidth]{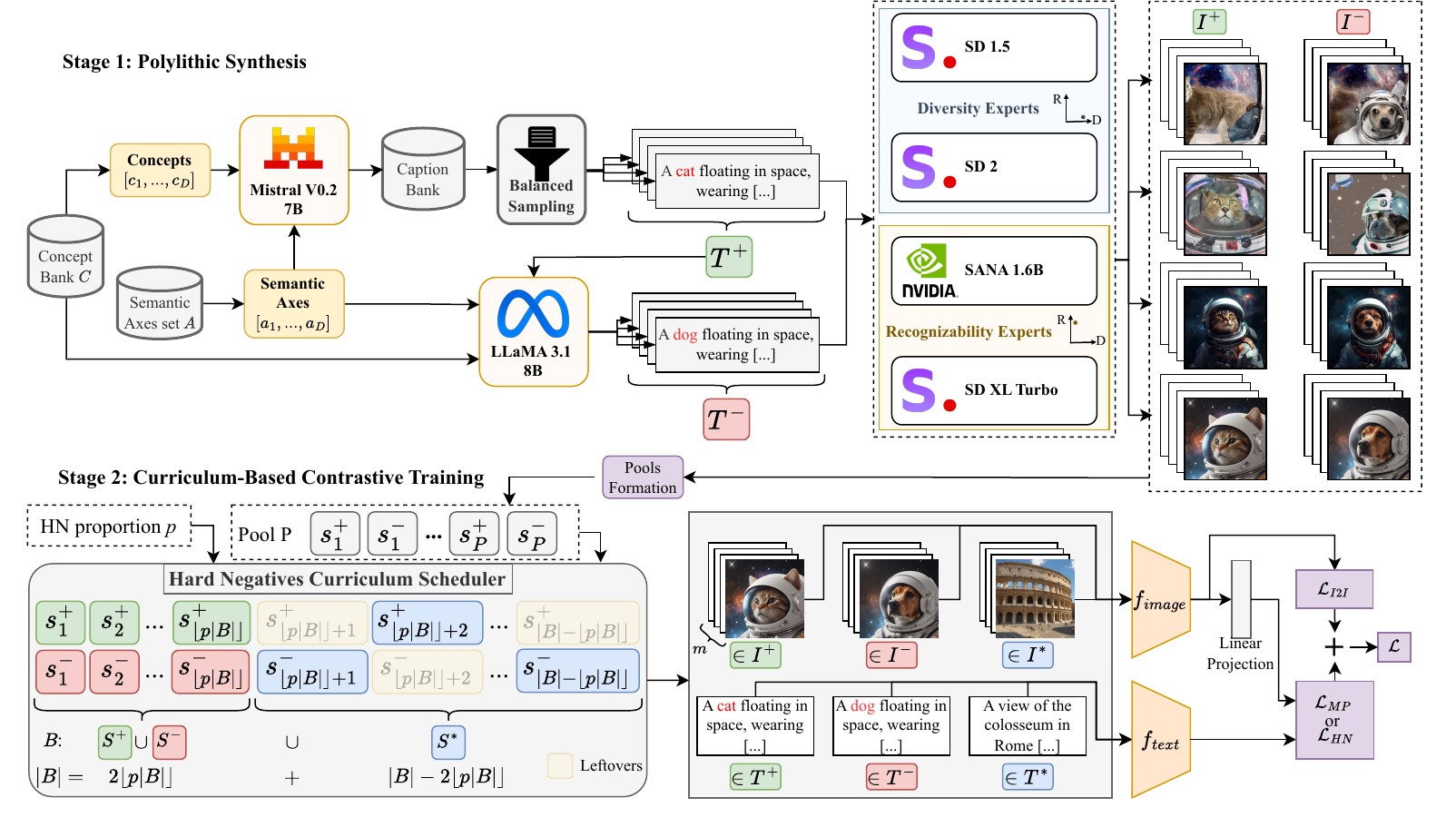}
    \vspace{-20pt}
    \caption{\textbf{PolyGen Pipeline Overview.} \textit{Stage 1 (Polylithic Synthesis):} Concepts from MetaCLIP Bank and semantic axes feed into a dual-LLM system. Mistral V0.2 generates base captions $T^+$ conditioned on concept-attribute pairs, while Llama 3.1 produces hard negatives $T^-$ by modifying specific semantic dimensions. Caption pairs are rendered through four diffusion models: Diversity Experts (SD 1.5, SD 2) ensure broad manifold coverage despite lower fidelity, while Recognizability Experts (SANA, SDXL-Turbo) provide high prompt adherence despite limited variance. \textit{Stage 2 (Curriculum-Based Training):} A scheduler linearly increases hard negative proportion $p$ from 0 to 0.5 during training. Unused samples populate a leftover queue, maintaining data efficiency. Batches combine multi-positive samples ($I^+, T^+$) and hard negatives ($I^-, T^-$) to compute $\mathcal{L}_{I2I}$ and $\mathcal{L}_{HN}$.}
    \label{fig:pipeline}
    \vspace{-10pt}
\end{figure*}

\section{Methodology}\label{sec:method}
We introduce \textbf{PolyGen} (Polylithic Generation Framework), an end-to-end pipeline for training vision-language models on fully synthetic data. The framework consists of two stages (Figure \ref{fig:pipeline}): \textit{Polylithic Synthesis} combines structured caption generation with multi-generator image rendering, followed by \textit{Curriculum-Based Contrastive Training}. We detail each component below.

\subsection{Structured Caption Pairs}
The first component of \textit{Polylithic Synthesis} (Stage 1, Figure \ref{fig:pipeline}) generates semantically controlled caption pairs.
Prior work on synthetic data generation typically relies on ``creative'' prompting, generating descriptive captions without semantic control. We instead adopt a constructive approach, generating training pairs $(T^+, T^-)$ through controlled semantic modifications.

\textbf{Attribute-Conditioned Caption Generation.}
We use Mistral-V0.2-7B \citep{jiang2023mistral7b} to generate the base captions $T^+$, conditioning the generation on a control tuple $(c \in \mathcal{C}, a \in \mathcal{A})$. Here, $c$ is the base \textit{Concept}, i.e., the primary subject sampled from the MetaCLIP Concept Bank \citep{xu2023demystifying}, and $a$ is the \textit{semantic modification axis} along which the hard negative will be generated. $\mathcal{A}$ includes visual attributes (\textit{Lighting, Material, Perspective, Color, Background, Position, Style}, ablated in Appendix \ref{app:axes_selection}) and an explicit \textit{Concept} axis for ``change of subject'' negatives. 
The additional attribute indication also serves to increase variability when scaling generation leads to the use of the same concept multiple times. (full generation prompts reported in Appendix \ref{app:prompts}). $T^+$ is obtained after downsampling a broader pool of captions using MetaCLIP's balanced sampling algorithm (Appendix~\ref{app:balanced_sampling}).

\begin{figure*}[t!]
    \centering
    \includegraphics[width=\textwidth]{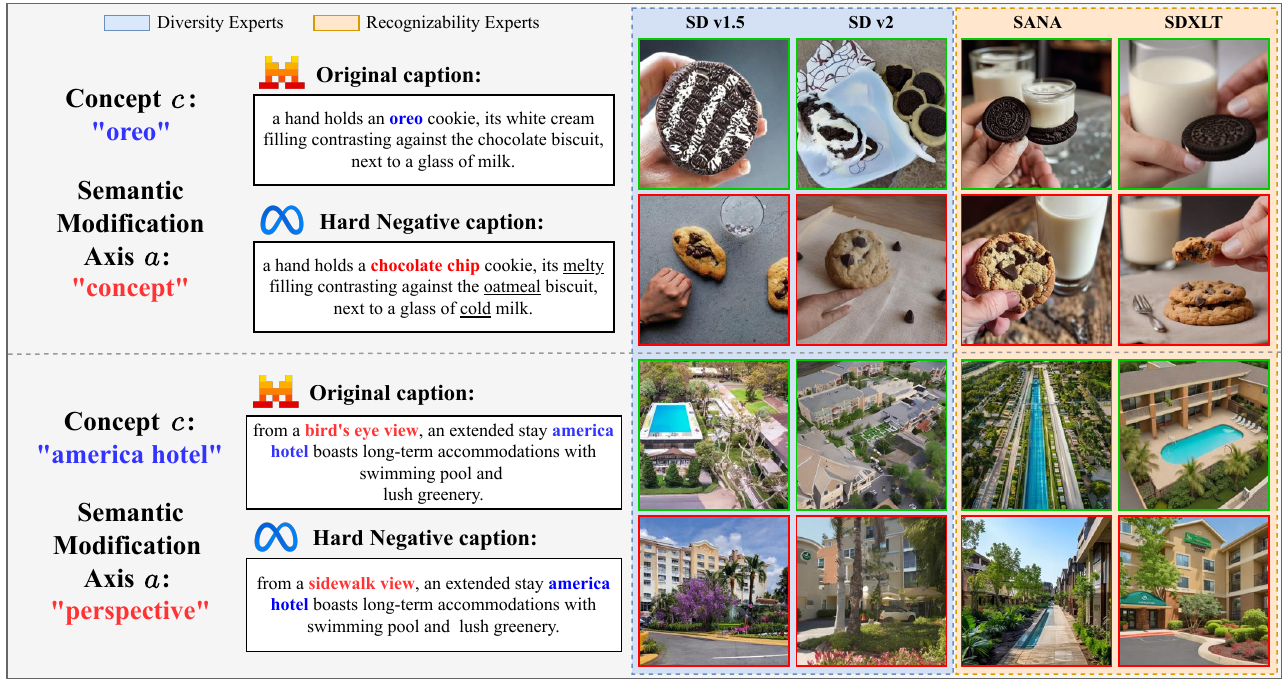}
    \vspace{-15pt}
    \caption{\textbf{Hard Negative Generation Across Multi-Generator Ensemble.} Two examples of controlled semantic perturbations rendered through four diffusion models. \textit{Top:} Concept-level modification (``oreo'' $\rightarrow$ ``chocolate chip cookie'', $a =$ Concept). \textit{Bottom:} Viewpoint transformation (``bird's eye view'' $\rightarrow$ ``sidewalk view'', $a =$ Perspective). Diversity Experts (SD 1.5, SD 2) produce high intra-concept variance with lower photorealism, while Recognizability Experts (SANA, SDXL-Turbo) generate photorealistic outputs with reduced stylistic variation. The architectural heterogeneity forces models to learn generator-invariant semantic representations, as the only consistent signal across columns is the underlying concept modification.}
    \label{fig:polygen_ex}
    \vspace{-10pt}
\end{figure*}

\textbf{Hard Negative Generation.} 
Thanks to the structured base captions, we can leverage LLMs to generate controlled negatives. We use Llama-3.1-8B \citep{dubey2024llama} (ablated in \cref{sec:llm_hn_ablation}) to generate the counterfactual hard negatives $T^-$. The model receives $t\in T^+$, the corresponding semantic axis $a$, and concept $c$, with strict instructions to only modify the specified axis while preserving syntactic structure. As illustrated in \Cref{fig:polygen_ex}, this strategy ensures global semantic coherence. Unlike rigid programmatic substitution, which risks generating logical absurdities by modifying terms in isolation, the LLM dynamically adjusts contextually dependent attributes to align with the new concept.

\subsection{Multi-Generator Synthesis Ensemble}
The second component of Stage 1 renders each caption pair through multiple diffusion models.
A core challenge in synthetic training is to determine the optimal generator $\mathcal{G}$. Relying on a single model risks learning a biased representation of the data manifold: each generator captures only a subset of the visual-semantic spectrum, introducing architecture-specific frequency patterns and compression artifacts that the downstream model may exploit as spurious features rather than learning robust semantic content.

\textbf{Metric-Driven Selection.} To systematically evaluate candidate generators, we establish a quantitative framework for measuring two complementary properties of synthetic datasets in open-vocabulary contrastive settings. While \citet{fan2024scaling} defined \textit{Recognizability} and \textit{Diversity} for supervised classification datasets (with fixed label sets), we adapt these metrics for the CLIP (self-supervised) paradigm:

\noindent \textit{- Recognizability (Fidelity).} In the absence of class labels, we measure the alignment between the generated image $x$ and its conditioning caption $t$. We define Recognizability as the expected CLIPScore \citep{hessel2021clipscore} over the dataset $\mathcal{D}$:
\begin{equation}
R(\mathcal{D}) = \frac{1}{|\mathcal{D}|}\sum_{(x,t)\in\mathcal{D}} \text{CLIPScore}(f_{img}(x), f_{txt}(t))
\end{equation}
High recognizability indicates the generator reliably respects the semantic constraints of the prompt.

\begin{figure}[t!]
    \centering
    \includegraphics[width=\linewidth]{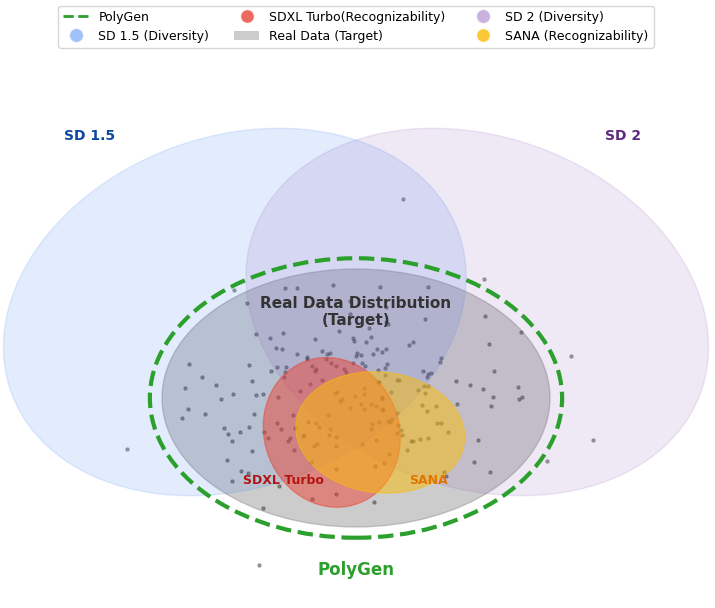} 
    \caption{\textbf{Conceptual Illustration of the Generator-Invariance Hypothesis.} Individual generators (colored circles) act as ``Experts'' with specific biases: older models like SD 1.5 (Blue) offer high diversity but low fidelity, while modern distilled models like SDXLT (Red) offer high fidelity but suffer from mode collapse. PolyGen (Green Dashed) trains on the union of these manifolds, approximating the coverage of the Real Data Distribution (Grey) by marginalizing out generator-specific artifacts.}
    \label{fig:invariance_hypothesis}
\end{figure}

\noindent \textit{- Diversity:} Measuring diversity in open-world data is challenging due to the high dimensionality of the visual space and the semantic relativity of what constitutes meaningful variation. We propose a cluster-based approach that captures intra-concept variance. We partition the dataset into $K=1000$ semantic clusters based on caption embeddings, then define Diversity as the average standard deviation of image embeddings within each cluster $k$:
\begin{equation}
D(\mathcal{D}) = \frac{1}{K}\sum_{k=1}^{K} \sqrt{\frac{1}{|\mathcal{D}_k|}\sum_{x \in \mathcal{D}_k} ||f_{img}(x) - \mu_k||^2}
\end{equation}
where $\mu_k$ is the centroid of image embeddings in cluster $k$. This metric specifically captures the ability of the generator to produce visually distinct interpretations of the same semantic concept, or group of concepts.

\textbf{The Generator-Invariance Strategy.} Guided by the \textit{Generator-Invariance Hypothesis} (\cref{fig:invariance_hypothesis}), we construct an ensemble $\mathcal{G}$ that maximizes coverage of the visual-semantic spectrum. We select four architecturally distinct models: 

- \textit{Recognizability Experts:} SDXL-Turbo \citep{sauer2024adversarial} and SANA-1.6B \citep{xie2024sana}, selected for high prompt adherence and photorealism despite lower output diversity. 

- \textit{Diversity Experts:} Stable Diffusion v1.5 and v2 \citep{ldm}, selected for high variance ensuring broad manifold coverage despite lower fidelity.

For every caption pair $(t^+, t^-)$, we generate corresponding images using all four models. This forces the downstream model to learn representations invariant to generator-specific spectral signatures such as the compression artifacts in SD1.5 or the low-frequency bias in SDXL-Turbo (Appendix~\ref{app:spectral_analysis}), as the only signal consistent across the ensemble is the semantic concept itself.

\subsection{Curriculum-Based Contrastive Training}
Stage 2 leverages the structured dataset through a specialized training pipeline that exploits the one-to-many mapping from captions to images (up to $n^+=4$ positive images each caption).

\textbf{Multi-Positive Objectives.} To exploit the ensemble of $n^+$ images generated for each caption $t_i$, we adopt a Multi-Positive objective \citep{tian2023stablerep}. Rather than one-hot targets, we use soft targets distributing probability mass uniformly ($1/n^+$) across all images from the same prompt. This objective $\mathcal{L}_{MP}$ aligns the text embedding with the semantic centroid of the diverse visual manifold formed by the generator ensemble. Beyond text-to-image alignment, we evaluate the impact of explicit Image-to-Image regularization. We incorporate the contrastive term $\mathcal{L}_{I2I}$ to penalize the model for learning generator-specific idiosyncrasies. For any image anchor $i$, we define a ground-truth distribution $p_i$ that is uniform ($1/(n^+-1)$) over the other images generated from the same caption. Given the predicted categorical distribution:

\begin{equation}
    q_{i,j} = \frac{ e^{(s(i,j)/\tau)}}{\sum_{k\in I\setminus \{i\}} e^{s(i,k)/\tau)}},
\end{equation}
the loss is computed as the cross-entropy:
\begin{equation}
    \mathcal{L}_{I2I} = - \frac{1}{N}\sum_{i \in I}\left( \sum_{j\in I\setminus \{i\}} p_{i,j} \log q_{i,j}\right).
\end{equation}
This term acts as a structural regularizer, forcing the visual encoder to discard ``spurious'' features such as spectral artifacts or stylistic biases, and move the representation toward the invariant semantic core defined by the prompt.

\noindent
\textbf{Hard Negatives Objectives.} Standard InfoNCE suppresses hard negatives $T^-$ by treating them as any other sample in the batch denominator. To enforce fine-grained discrimination, we employ the TripletCLIP loss \citep{patel2024tripletclip}. 

We define the Hard-Negative NCE loss for image-to-text as:
\begin{equation}
    \mathcal{L}^{I^+\rightarrow T^+;T^-}_{NCE} = -\frac{1}{N} \sum_{k=1}^N \log \frac{e^{s(i_k, t_i)/\tau}}{\sum_{t\in T^+\cup T^-} e^{s(i_k, t)/\tau}},
\end{equation}
analogous for text-to-image. The NegCLIP loss is then:
\begin{equation}
    \mathcal{L}^{I^+, T^+, T^-}_{NegCLIP} = \mathcal{L}^{I^+ \rightarrow T^+;T^-}_{NCE} + \mathcal{L}^{T^+ \rightarrow I^+}_{NCE}.
\end{equation}
Finally, the total TripletCLIP loss is defined as:
\begin{equation}
    \mathcal{L}_{TripletCLIP} = \mathcal{L}^{I^+, T^+, T^-}_{NegCLIP}  + \mathcal{L}^{I^-, T^-, T^+}_{NegCLIP}.
\end{equation}
This formulation forces the model to explicitly distinguish the base concept from its semantic counterfactual.

\textbf{Hard Negatives Curriculum Scheduler.}
Naive injection of hard negatives causes early training instability because fine-grained verification tasks disrupt coarse-grained semantic alignment, preventing stable conceptual clusters.
We introduce a curriculum scheduler linearly increasing the in-batch hard negative ratio $p$ from $0 \to 0.5$ over the training epochs.
To ensure full data utilization throughout the epoch, we implement a \textit{Leftover Queue}: samples that are not paired with their corresponding hard negatives are queued and batched in subsequent steps. 
This scheduler, when used in combination with a decaying learning rate, allows the model to learn a more robust knowledge baseline first, and exploits hard negatives for learning fine-grained discriminations later in training.
Since batch composition varies, we adapt $\mathcal{L}_{HN}$ to balance subset contributions:
\begin{align}
    \mathcal{L}_{HN} =& \frac{1}{|B|}(|T|\mathcal{L}_{NegCLIP}^{I, T, T^-}+ |T^-|\mathcal{L}_{NegCLIP}^{I^-, T^-, T}),
\end{align}
with $T = T^*\cup T^+$, $I = I^*\cup I^+$, and $|B| = |T|+|T^-|$.

\section{Experiments and Results}\label{sec:experiments}

\subsection{Experimental Setup}
\textbf{Training Configuration.} To ensure a rigorous comparison with prior state-of-the-art SynthCLIP \citep{hammoud2024synthclip}, we conduct our primary evaluation on a controlled subset of 500k captions sampled from the CC3M dataset.
We train a ViT-B/16 CLIP architecture from scratch for 40 epochs. We use the AdamW optimizer with a learning rate of $5 \times 10^{-4}$ and a cosine decay schedule with 1 warmup epoch. For PolyGen, we use a batch size of 1536. To handle the Multi-Positive input fairly, we fix the number of training images and scale the number of captions accordingly: this corresponds to $500k/n^+$ captions in case of $n^+$ positives.\\  The Hard Negative ratio $p$ follows our curriculum scheduler, linearly increasing from 0 to 0.5 over the training course.

\textbf{Downstream Tasks}
We adopt SynthCLIP evaluation setting, evaluating our models in both vision-only tasks, meaning the quality of the features extracted by the image encoder (i.e., linear probing, few-shot classification), and vision-language tasks, with zero-shot retrieval and classification.
Also following SynthCLIP, results are summarized into a unique $\Delta_{MTL}$ metric \cite{deltamtl} (Appendix~\ref{app:deltamtl}), aggregating top-1 accuracy for vision tasks, top-5 accuracy for zero-shot classification, and recall@5 for retrieval, representing the relative performance drop or improvement w.r.t. a baseline model, while we report full per-task results in Appendix~\ref{app:results}.\\

\textbf{Datasets.}
The models are evaluated on a comprehensive suite of downstream benchmarks to assess both vision-only and vision-language capabilities:
for Linear Probing \& Few-Shot we use: CIFAR10 and CIFAR100 \cite{cifar}, FGVC Aircraft \cite{aircraft}, DTD (Textures) \cite{dtd}, Oxford Flowers \cite{flowers}, Oxford-IIIT Pets \cite{pets}, SUN397 \cite{sun}, Caltech-101 \cite{caltech101}, Food-101 \cite{food101}.
We also use ImageNet \cite{imagenet} for zero-shot classification (accuracy), Flickr8k and Flickr30k \cite{flickr8k}, and MS COCO \cite{coco} for text-to-image and image-to-text retrieval. Lastly, for hard negative tasks, we use SugarCrepe \cite{hsieh2023sugarcrepe} and SugarCrepe++ \cite{dumpala2024sugarcrepe++}. Results in this section are reported per-task, as averages over the datasets. Per-dataset results are reported in Appendix~\ref{app:results}.

\begin{figure}[t!]
    \centering
    \includegraphics[width=\columnwidth]{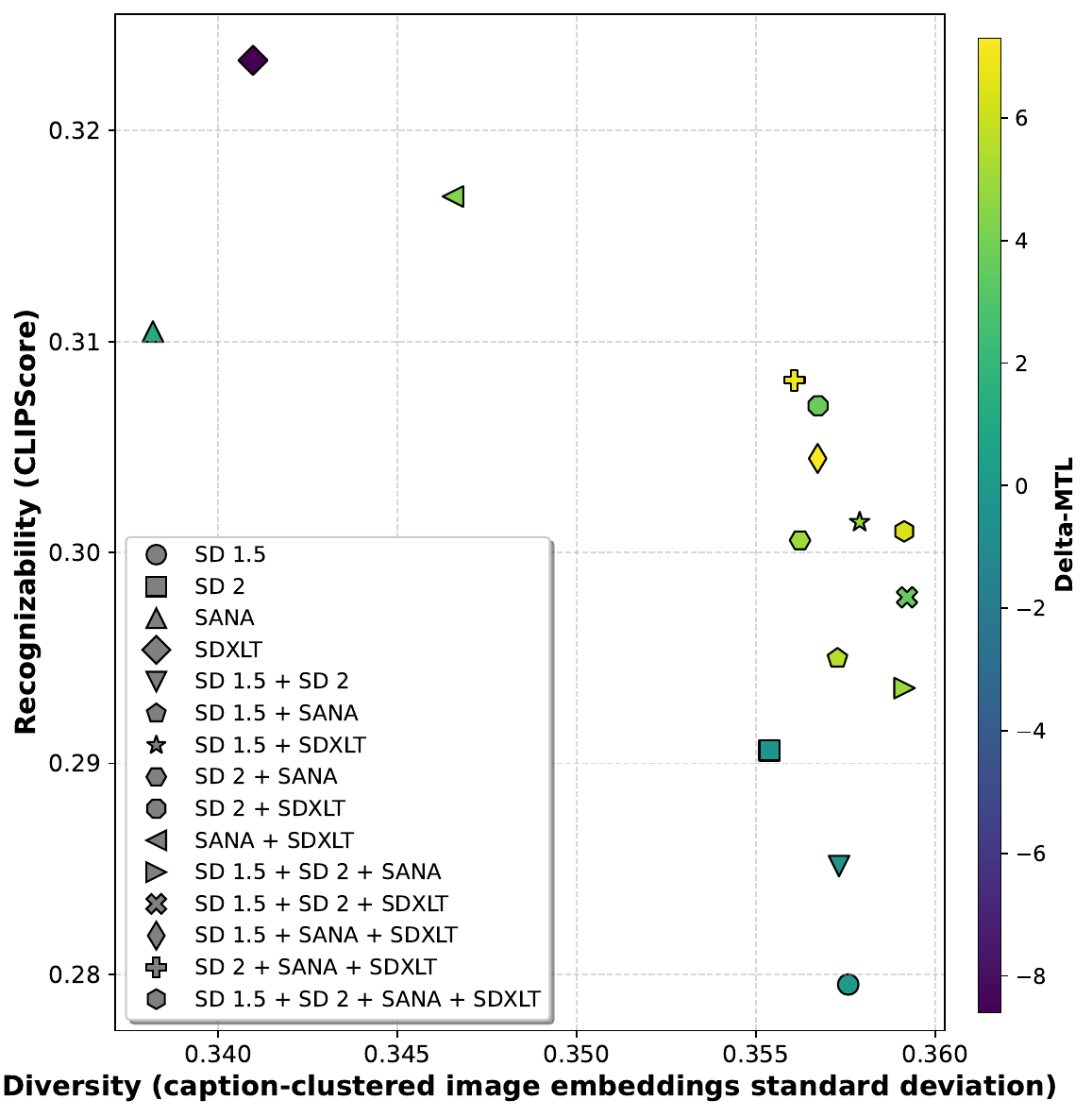}
    \vspace{-10pt}
    \caption{Recognizability vs. Diversity, with color indicating downstream performance. While individual models trade off diversity or recognizability, PolyGen ensembles break this Pareto frontier,  maximizing both axes to achieve superior $\Delta_{MTL}$.}
    \label{fig:rec_vs_div}
    \vspace{-15pt}
\end{figure}

\subsection{Recognizability vs. Diversity implications}
\Cref{fig:rec_vs_div} illustrates the relationship between the distributional properties of the training data (Recognizability vs. Diversity) and the downstream performance (indicated by color). 
To isolate the impact of manifold coverage, we train a series of CLIP models on fixed sets of 500k image-text pairs ($n^+=1$). For the ensemble runs ($m>1$), the image for each caption is uniformly sampled from the pool of selected generators.
This ensures that any performance gain is attributed to inter-source diversity rather than an increase in the number of training samples. 
The plot reveals a clear dichotomy in generative behaviors. Older architectures, specifically Stable Diffusion (SD) v1.5, occupy the high-diversity, lower-recognizability region. In contrast, modern distilled models like SDXL-Turbo and SANA-1.6B exhibit significantly higher Recognizability but at a noticeable cost to intra-concept Diversity. This phenomenon is representative of ``mode collapse'' in high-fidelity generators, where the model converges on a limited set of high-probability visual interpretations to ensure aesthetic appeal. Our strategy effectively moves the training distribution toward the upper-right quadrant, effectively improving performance on downstream tasks. This empirically validates our Generator-Invariance Hypothesis: while individual generators are limited by their specific spectral biases, the ensemble marginalizes these artifacts. By forcing the model to learn features consistent across disjoint manifolds, we achieve robust generalization ($\Delta_{MTL}$) that neither type of generator could achieve in isolation.

\begin{table}[t!]
    \centering
    \resizebox{\columnwidth}{!}{
        \renewcommand{\arraystretch}{1.4}
        \begin{tabular}{cc|c|ccccc|cc}
        \cline{1-10}
        & & \multirow{2}{*}{\rotatebox{90}{\parbox{2.3cm}{\centering $\mathcal{L}_{I2I}$}}} & \multirow{2}{*}{\rotatebox{90}{\parbox{2.3cm}{\centering \textit{L.P.}}}} & \multirow{2}{*}{\rotatebox{90}{\parbox{2cm}{\centering \textit{Few-shot}}}} & \multirow{2}{*}{\rotatebox{90}{\parbox{2cm}{\centering \textit{Img Ret.}}}} & \multirow{2}{*}{\rotatebox{90}{\parbox{2cm}{\centering \textit{Text Ret.}}}} & \multirow{2}{*}{\rotatebox{90}{\parbox{2cm}{\centering \textit{0-shot}}}} & \multicolumn{2}{c}{$\Delta_\text{MTL}$ vs.} \\[0.6cm]

        $n^+$ & $m$ & & & & & & &
        SynthCLIP & \makecell{same $n^+$,\\ $m=1$}\\

        \cline{1-10}

        1 & 1 & {\color{pred}{\xmark}}& 49.1 & 61.6 & 11.6 & 15.9 & 8.15 & - & - \\

        1 & 2 & {\color{pred}{\xmark}}& 47.5 & 61.1 & 12.1 & 17.5 & 8.53 & {\color{pgreen}{+3.1\%}} & {\color{pgreen}{+3.1\%}}\\
        1 & 3 & {\color{pred}{\xmark}}& 48.5 & 61.3 & 13.3 & 17.9 & 8.56 & {\color{pgreen}{+6.1\%}} & {\color{pgreen}{+6.1\%}} \\

        1 & 4 & {\color{pred}{\xmark}}& 48.9 & 61.5 & 12.8 & 17.6 & 8.92 & {\color{pgreen}{+6.2\%}} & {\color{pgreen}{+6.2\%}}\\
        \cdashline{1-10}

        2 & 1 & {\color{pred}{\xmark}}& 49.5 & 62.4 & 13.8 & 18.7 & 9.41 & {\color{pgreen}{+11.0\%}} & -\\

        2 & 2 & {\color{pred}{\xmark}}& 47.9 & 61.1 & 14.2 & 20.2 & 9.54 & {\color{pgreen}{+12.7\%}} & {\color{pgreen}{+1.3\%}} \\

        3 & 1 & {\color{pred}{\xmark}}& 49.3 & 62.7 & 14.7 & 21.1 & 9.66 & {\color{pgreen}{+16.1\%}} & -\\

        3 & 3 & {\color{pred}{\xmark}}& 48.2 & 61.1 & 16.2 & \textbf{22.4} & 9.38 & {\color{pgreen}{+18.7\%}} & {\color{pgreen}{+1.7\%}} \\

        4 & 1 & {\color{pred}{\xmark}}& \textbf{49.8} & \textbf{62.8} & 14.8 & 19.4 & 9.56 & {\color{pgreen}{+14.1\%}} & -\\

        4 & 4 & {\color{pred}{\xmark}}& 48.9 & 62.0 & 16.1 & 21.4 & 9.86 & {\color{pgreen}{\textbf{+19.0\%}}} & {\color{pgreen}{+3.8\%}}\\

        \cdashline{1-10}

        2 & 1 & {\color{pgreen}{\checkmark}}& 43.4 & 57.3 & 13.3 & 17.6 & 10.14 & {\color{pgreen}{+6.3\%}} & -\\

        2 & 2& {\color{pgreen}{\checkmark}} & 41.5 & 55.7 & 14.4 & 19.8 & 11.19 & {\color{pgreen}{+12.3\%}} & {\color{pgreen}{+4.8\%}}\\

        3 & 1 & {\color{pgreen}{\checkmark}}& 44.4 & 57.3 & 14.2 & 18.9 & 10.88 & {\color{pgreen}{+11.8\%}} & -\\

        3 & 3& {\color{pgreen}{\checkmark}} & 41.3 & 55.2 & \textbf{16.5} & 21.7 & 10.98 & {\color{pgreen}{+17.4\%}} & {\color{pgreen}{+4.1\%}}\\

        4 & 1& {\color{pgreen}{\checkmark}} & 45.2 & 57.7 & 15.0 & 19.9 & 11.24 & {\color{pgreen}{+15.6\%}} & -\\
 
        4 & 4& {\color{pgreen}{\checkmark}} & 42.9 & 56.4 & 15.7 & 20.6 & \textbf{11.34} & {\color{pgreen}{+16.6\%}} & {\color{pgreen}{+0.4 \%}}\\
        \hline
        \end{tabular}
    }
    \vspace{5pt}
    \caption{Heterogeneity ($m$) vs. Density ($n^+$) averaged results. Jointly scaling $m$ and $n^+$ maximizes $\Delta_{MTL}$. Notably, strong invariance ($m=4, \mathcal{L}_{I2I}$) improves abstract reasoning (Zero-shot) at the cost of texture-based classification (L.P. : Linear Probing).}
    \label{tab:benchmark}
    \vspace{-15pt}
\end{table}

\subsection{The Heterogeneity-Density Trade-off}
\label{sec:main_results}

We isolate the impact of our two core design variables: the ensemble size ($m$) and the positive sample density ($n^+$). Table \ref{tab:benchmark} presents the results, starting from $m=1$ (SD 1.5), $m=2$ including SDXLT, $m=3$ also including SANA, and with $m=4$ indicating the full PolyGen ensemble.

\textbf{The Heterogeneity Gain ($m \uparrow$).}
Increasing the ensemble size from single-source ($m=1$) to the full ensemble ($m=4$) yields a universal improvement in $\Delta_{MTL}$ across all configurations.
Critically, this gain is driven by Vision-Language tasks. For example, in the $n^+=1$ block, 0-shot accuracy rises from 8.15 ($m=1$) to 8.92 ($m=4$). This confirms the Generator-Invariance Hypothesis: seeing the same concept rendered by diverse engines forces the model to decouple semantics from generator-specific artifacts, resulting in more robust open-world representations.

\textbf{The Density Gain ($n^+ \uparrow$).}
While diversity optimizes generalization, density optimizes robustness. Increasing the number of positive pairs ($n^+$) improves performance on both axes (Vision-Only and Vision-Language).
By clustering multiple images around a single caption, the model learns a stable ``centroid'' representation. The combination of both factors is multiplicative: the full PolyGen configuration ($n^+=4, m=4$) reaches a peak $\Delta_{MTL}$ of +19.0\%, significantly outperforming methods that rely on density alone ($n^+=4, m=1$, +14.1\%).

\textbf{Invariance as a Regularizer.}
A revealing pattern emerges when analyzing the impact of explicit invariance mechanisms: the Image-to-Image loss ($\mathcal{L}_{I \to I}$) and the Ensemble size ($m$).
While increasing $m$ or adding $\mathcal{L}_{I \to I}$ maximizes Zero-Shot Accuracy (with the latter achieving the table's peak), it induces a regression in Linear Probing. For instance, single-source models often retain slightly higher probing scores than their ensemble counterparts ($m=4$).
We attribute this to the ``Semantic Purist'' nature of our framework. Simple linear classifiers often exploit low-level texture shortcuts or generator-specific spectral artifacts to discriminate classes. 

By enforcing invariance across diverse manifolds (via multi-positives) or explicitly penalizing feature divergence (via $\mathcal{L}_{I \to I}$), PolyGen strips away these non-semantic features. While this removes the textural cues used by linear probes, it forces the model to learn robust, abstract representations that transfer better to open-world tasks.
Ultimately, even with this trade-off, the full PolyGen configuration yields the highest $\Delta_{MTL}$ improvement, offering a pareto-optimal solution that prioritizes semantic alignment, the primary goal of Vision-Language pre-training, while maintaining competitive visual representation.

\subsection{Hard Negatives and Curriculum}
Table \ref{tab:hn_benchmark} provides a granular dissection of the training dynamics, particularly within the $m=2, n^+=2$ block.

\textbf{The Limits of Passive and Delayed Learning.}
We first establish the baseline behavior using ``Naive'' Hard Negatives, simply including counterfactual captions in the batch without targeted objectives. This approach yields high compositional performance (SugarCrepe++ 40.0\%) because the model is exposed to fine-grained variations throughout training. However, this early complexity acts as a distracter for general semantic alignment, resulting in a modest improvement on standard tasks ($\Delta_{MTL}$ +4.7\%) compared to the cleaner runs in the previous section.
Attempting to solve this via the Curriculum Scheduler alone introduces the inverse problem. By delaying the introduction of hard negatives, we shield the early training phase, allowing standard tasks to improve ($\Delta_{MTL}$ +7.0\%). Yet, this ``delayed viewing'' comes at the cost of compositional reasoning: the model simply does not see the difficult counterfactuals often enough to learn the subtle syntax of attribute binding.

\textbf{Impact of the Hard Negative loss.}
The behavior of $\mathcal{L}_{HN}$ in isolation proves effectively, with a higher $\Delta_{MTL}$ compared to the same configuration using $\mathcal{L}_{MP}$.
We attribute this to a key change of TripletCLIP loss, to which $\mathcal{L}_{HN}$ is based, and standard CLIP.

\begin{table}[t!]
    \centering
    \resizebox{\columnwidth}{!}{
        \renewcommand{\arraystretch}{1.4}
        \begin{tabular}{cc|c|c|c|ccccc|c|cc|c}
        \hline
        $n^+$ & $m$ & \rotatebox{90}{\parbox{2cm}{\centering \textit{HN samples}}} & \rotatebox{90}{\parbox{1cm}{\centering $\mathcal{L}_{HN}$}} & \rotatebox{90}{\parbox{1.6cm}{\centering \textit{scheduler}}} & \rotatebox{90}{\parbox{1.8cm}{\centering \textit{L.P.}}} & \rotatebox{90}{\parbox{1.8cm}{\centering \textit{Few-shot}}} & \rotatebox{90}{\parbox{1.8cm}{\centering \textit{Img Ret.}}} & \rotatebox{90}{\parbox{1.8cm}{\centering \textit{Text Ret.}}} & \rotatebox{90}{\parbox{1.8cm}{\centering \textit{0-shot}}} & $\Delta_\text{MTL}$ &  \rotatebox{90}{\parbox{2.3cm}{\centering SugarCrepe}} &  \rotatebox{90}{\parbox{2.3cm}{\centering SugarCrepe++}} & $\Delta^{HN}_\text{MTL}$  \\
        \hline
        1 & 1 & {\color{pred}{\xmark}} & {\color{pred}{\xmark}} & {\color{pred}{\xmark}} & \textbf{49.1} & \textbf{61.6} & 11.6 & 15.9 & 8.15 & - & 55.4 & 36.7 & - \\
        \cdashline{1-14}
        2 & 2 & {\color{pgreen}{\checkmark}}& {\color{pred}{\xmark}} & {\color{pred}{\xmark}} & 41.5 & 55.7 & 13.2 & 17.9 & 9.95 & {\color{pgreen}{+4.7\%}} &56.9&40.0&{\color{pgreen}{+5.9\%}}\\
        2 & 2 & {\color{pgreen}{\checkmark}}& {\color{pred}{\xmark}} & {\color{pgreen}{\checkmark}} & 40.8 & 55.8 & 13.6 & 18.8 & 10.23 & {\color{pgreen}{+7.0\%}} &55.0&38.8&{\color{pgreen}{+2.6\%}}\\
        2 & 2 & {\color{pgreen}{\checkmark}}& {\color{pgreen}{\checkmark}} & {\color{pred}{\xmark}} & 42.3 & 56.8 & 14.0 & 19.0 & 10.27 & {\color{pgreen}{+9.0\%}} &55.8&39.2&{\color{pgreen}{+3.7\%}}\\
        2 & 2 & {\color{pgreen}{\checkmark}}& {\color{pgreen}{\checkmark}} & {\color{pgreen}{\checkmark}} & 41.4 & 55.8 & 14.6 & \textbf{20.5} & 10.8 & {\color{pgreen}{+12.5\%}}&56.4&40.5&{\color{pgreen}{+6.1\%}}\\
        \cdashline{1-14}
        3 & 3 & {\color{pgreen}{\checkmark}}& {\color{pgreen}{\checkmark}} & {\color{pgreen}{\checkmark}} & 40.8 & 55.5 & \textbf{15.7} & 18.9 & 10.72 & {\color{pgreen}{+11.8\%}}&\textbf{58.1}&\textbf{41.6}&{\textbf{\color{pgreen}{+9.1\%}}}\\
        4 & 4 & {\color{pgreen}{\checkmark}} & {\color{pgreen}{\checkmark}} & {\color{pgreen}{\checkmark}} & 42.7 & 56.5 & \textbf{15.7} & 20.4 & \textbf{11.1} & {\color{pgreen}{\textbf{+15.5\%}}}&56.8&40.7&{\color{pgreen}{+6.8\%}}\\
        \hline
        \end{tabular}
    }
    \vspace{5pt}
    \caption{Average per-task results of models using hard negatives. From the second to the last row, models use $\mathcal{L}_{I2I}$. When used, $\mathcal{L}_{HN}$ replaces $\mathcal{L}_{MP}$.}
    \label{tab:hn_benchmark}
    \vspace{-20pt}
\end{table}

As the generated images often fail to correctly represent slight variations in captions (Appendix~\ref{app:t2i_hn_class}), TripletCLIP removes from the loss $T^+\rightarrow I^-$ and $T^-\rightarrow I^+$ comparisons. This allows the model to keep exploiting hard negatives ($I^+\rightarrow T^-$ and $I^-\rightarrow T^+$), while not being misdirected by the noisier signal given by hard negative images. The relation between CLIP and TripletCLIP loss is clearly shown in Appendix~\ref{app:tripletclip}.
On the other hand, removing such components of the loss leads to sub-optimal performance gains in compositional benchmarks.

\textbf{Optimization Synergy.}
The full PolyGen configuration empirically resolves the stability-plasticity dilemma. By combining the Scheduler and Triplet Loss, we achieve a $\Delta_{MTL}$ of +12.5\%, surpassing both the Scheduler-only and Loss-only configurations. This indicates a corrective mechanism: the Scheduler protects early feature formation from noisy gradients, while the Triplet Loss compensates for the reduced exposure time by enforcing a stricter margin on the hard negatives once they are introduced. This recovery allows us to match the high compositional score of the Naive approach without sacrificing general performance.

\textbf{Scaling and the Pareto Frontier.}
Consistent with our main results, scaling the ensemble size ($m$) and positive density ($n^+$) yields monotonic gains for general open-world tasks, with the $m=4$ configuration reaching the $\Delta_{MTL}$ peak.
However, for compositional reasoning, we observe a distinct peak at $m=3$, outperforming $m=4$.
We attribute this to a \textit{variance saturation point}: while maximal diversity ($m=4$) benefits broad semantic coverage, it introduces excessive visual noise for fine-grained syntactic tasks, making it harder for the model to isolate subtle attribute changes amidst drastic stylistic shifts.
Ultimately, the configuration $m=4$ represents the \textit{Pareto-optimal} choice: it maximizes general transferability while significantly improving compositional reasoning, offering the most robust balance between semantic breadth and syntactic depth.

\subsection{Ablation: LLM Selection for Semantic Synthesis}
\label{sec:llm_hn_ablation}
Our pipeline employs two different LLMs. For base captions, we use Mistral-v0.2-7B to maintain parity with the SynthCLIP baseline. Hard negative synthesis is, instead, more critical, as it demands high-fidelity instruction following to modify specific semantic attributes without altering the overall scene structure. To this end, we evaluated a suite of 7B/8B-parameter LLMs for correct negative generation: given a positive-negative caption pair generated by a model from a pair $(c,a)$, we verify if an LLM ensemble can recover the correct $a$ from the caption pair. As shown in Table \ref{tab:llm_selection}, Llama-3.1-8B significantly outperforms its peers, achieving 74\% classification accuracy on its generated negatives.

\begin{table}[t!]
\setlength{\tabcolsep}{3pt} 
\centering
\small
\begin{tabular}{l | c c c c}
\toprule
\textbf{Model} & \textbf{Accuracy} & \textbf{Avg. F1} & \textbf{Avg. Prec.} & \textbf{Avg. Rec.} \\
\midrule
Mistral V0.2 & 61\% & 60\% & 76\% & 61\% \\
Mistral V0.3 & 62\% & 62\% & 77\% & 62\% \\
Qwen 2.5 & 67\% & 67\% & 78\% & 67\% \\
Falcon 7B & 63\% & 65\% & 79\% & 63\% \\
\textbf{Llama 3.1} & \textbf{74\%} & \textbf{75\%} & \textbf{80\%} & \textbf{74\%} \\
\bottomrule
\end{tabular}
\vspace{5pt}
\caption{\textbf{LLM Selection for Hard Negatives.} Semantic Axes aggregated results of the LLM negative generation evaluation.}
\label{tab:llm_selection}
\vspace{-20pt}
\end{table}

\section{Discussion and Limitations}\label{sec:discussion}
\textit{The Synergy of Diversity and Recognizability.} 
Our results highlight a fundamental trade-off in current generative backbones when used for synthetic training VL pipelines: high-diversity models introduce noise that dilutes the learning signal, while high-fidelity models induce overfitting to narrow visual modes.
PolyGen resolves this by integrating these complementary manifolds. The high-fidelity generators provide strong "semantic anchors" that ensure the text concept is correctly instantiated, while the high-diversity generators expand the visual support of that concept, preventing the model from latching onto specific patterns. This synergy forces the model to ignore the distinct spectral fingerprints of each architecture and converge on the only shared signal: the invariant semantic content.

\textit{Why Synthetic Training?} 
The transition to synthetic pre-training is motivated by the need for curation over collection. Unlike web-scraping, synthetic pipelines offer precise control, enabling privacy-preserving, bias-aware, and multimodally scalable datasets. However, adoption has been hindered by the lower efficience of synthetic data. PolyGen contributes to take a step forward: a relevant factor for this "inefficiency" is single-generator homogeneity. By showing that manifold diversity drastically improves per-sample utility, we establish that synthetic pre-training is not just about scale, but it can be optimized for efficiency through structured ensemble sampling.

\textbf{Limitations and Future Work.}
Our approach entails specific design trade-offs. First, while not necessarily computationally more expensive, maintaining a heterogeneous ensemble introduces higher pipeline complexity than single-model baselines. Second, to adhere to a fixed budget, our multi-positive strategy prioritizes semantic robustness over unique concept coverage, a balance that requires recalibration for larger-scale regimes. Lastly, while our current concept bank and semantic axes ensure broad coverage, it remains a simplification of real-world visual complexity. Future work will extend the strategy along three dimensions: (1) increasing ensemble size, thus increasing both $m$ and $n^+$, with architectures showing diverse Recognizability and Diversity properties; (2) scaling the data budget to evaluate the breadth-depth Pareto frontier at larger-scale regimes; (3) expanding programmatic hard negatives to cover complex axes such as multi-concept relationships, counting, and causality, ultimately obtaining a synthetic curriculum capable of fully grounding open-world reasoning.

\section{Impact Statement}
This paper presents work whose goal is to advance the field of deep learning, specifically by enabling safer and more controllable vision-language pre-training via synthetic data. A primary motivation for our approach is to reduce reliance on large-scale web scraping, which inherently risks including private, non-consensual, or toxic content.

To mitigate the propagation of harmful biases, our pipeline leverages standard generative backbones (e.g., Llama 3, Stable Diffusion, SANA) that have undergone safety alignment and data filtering. Furthermore, our text generation is grounded in the MetaCLIP concept bank, a curated metadata set designed to provide a balanced and cleaner semantic distribution compared to raw web text. While we acknowledge that synthetic generators may still reflect the biases of their underlying training data, our framework offers a pathway to explicitly filter these biases programmatically. Given the controlled, methodological nature of this study (conducted as a proof-of-concept on a limited data budget), we do not foresee immediate negative societal consequences.

\section*{Acknowledgements}
This paper is supported by the FAIR (Future Artificial Intelligence Research) project, funded by the NextGenerationEU program within the PNRR-PE-AI scheme (M4C2, investment 1.3, line on Artificial Intelligence). We also acknowledge CINECA for providing the computational resources necessary to conduct the experiments and training for this work.

\bibliographystyle{plainnat}  
\bibliography{references}

\clearpage
\onecolumn
\appendix 

\section{Caption Generation Prompts}
\label{app:prompts}

\begin{figure}[t]
    \centering
    \setlength{\tabcolsep}{0pt}
    \subfloat[Prompt used by Mistral for the generation of the baseline caption set. Red text is only included when an attribute indication is added.\label{fig:prompt}]{\includegraphics[width=0.475\textwidth]{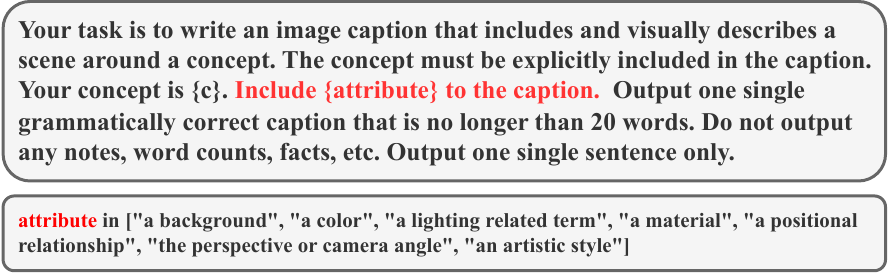}}
    \hspace{10pt}
    \subfloat[Prompt used by Llama for the generation of the hard negative caption set. When the original caption is generated without using any attribute, $c$ is used as the semantic axis to change.\label{fig:hn_prompt}]{\includegraphics[width=0.475\textwidth]{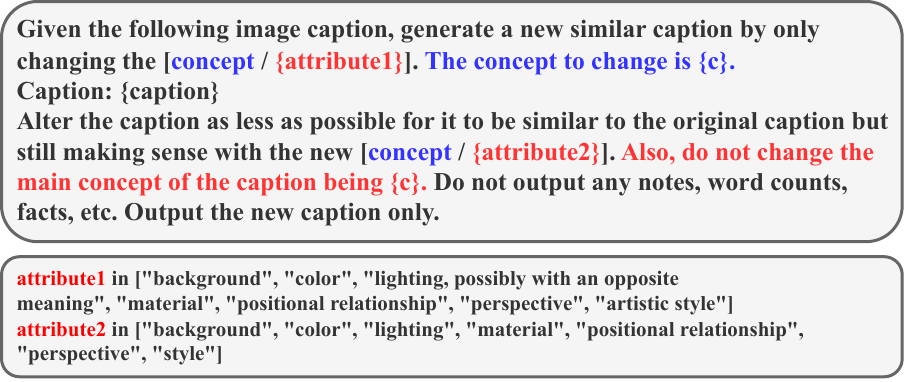}}
    \caption{\textbf{Zero-shot programmatic conditioning.} (a) Baseline prompt enforcing structured concept-attribute tuples $(c, a)$ to ensure semantic orthogonality. (b) Hard Negative prompt imposing strict counterfactual constraints to generate valid $T^-$ samples without syntactic drift.}
    \label{fig:prompts}
\end{figure}

\cref{fig:prompts} shows the zero-shot prompts we use to generate captions. The baseline set of captions is generated with the prompt shown in \cref{fig:prompt}, asking the model to generate a caption given the starting concept $c\in C$, and optionally, asks it to include an attribute $a$ from a predefined list of semantic modification axes $\mathcal{A}$. Hard negative captions are then generated with the prompt in \cref{fig:hn_prompt}, which instructs the LLM to modify the base caption on its selected semantic axis only, while maintaining the rest of the sentence structure.

\section{Relation between CLIP and TripletCLIP loss}
\label{app:tripletclip}

In this section, we provide the formal derivation connecting the standard CLIP objective with the TripletCLIP loss used in our Hard Negative mining strategy. We explicitly show how the Triplet formulation refines the contrastive landscape by filtering out noisy text-to-image negatives while maintaining a strict margin on the informative image-to-text pairs.

Given $T^+ = \{t^+_1, t^+_2, \ldots, t^+_N\}$, the set of base text features extracted from a text encoder:
\begin{equation}
    t^+_k = f_{txt}(\text{caption}_k), \quad \forall k \in [1, N],
\end{equation}
and $T^- = \{t^-_1, t^-_2, \ldots, t^-_N\}$ the set of hard negative text features extracted from $f_{txt}$ using the LLM-generated hard negative caption:
\begin{equation}
    \text{caption}_k^- = \text{LLM}(\text{caption}_k, \text{hard negative prompt}).
\end{equation}

Similarly, let $I^+ = \{i^+_1, i^+_2, \ldots, i^+_N\}$ be the set of base image features extracted from an image encoder:
\begin{equation}
    i^+_k = f_{img}(\text{image}_k), \quad \forall k \in [1, N],
\end{equation}
and $I^- = \{i^-_1, i^-_2, \ldots, i^-_N\}$ the set of hard negative image features extracted from $f_{img}$ using the diffusion-generated image:
\begin{equation}
    \text{image}_k^- = \text{T2I}(\text{caption}_k^-).
\end{equation}

We define $\tilde{T} = T^+ \cup T^-$ and $\tilde{I} = I^+ \cup I^-$ as the concatenated batches of all textual and visual features, respectively, with size $2N$.
The standard CLIP loss on this concatenated batch is defined as:
\begin{equation}
    \mathcal{L}_{CLIP}(\tilde{I}, \tilde{T}) = \frac{1}{2}(\mathcal{L}_{NCE}^{\tilde{I} \rightarrow \tilde{T}} + \mathcal{L}_{NCE}^{\tilde{T} \rightarrow \tilde{I}}).
\end{equation}

We can expand the Image-to-Text term $\mathcal{L}_{NCE}^{\tilde{I} \rightarrow \tilde{T}}$ by splitting the summation over the $2N$ samples into the positive anchors ($I^+$) and negative anchors ($I^-$):

\begin{align}
    \mathcal{L}_{NCE}^{\tilde{I} \rightarrow \tilde{T}} &= -\frac{1}{2N} \sum_{k=1}^{2N} \log \frac{\exp(s(\tilde{i}_k,\tilde{t}_k)/\tau)}{\sum_{j=1}^{2N}\exp(s(\tilde{i}_k,\tilde{t}_j)/\tau)} \\
    &= -\frac{1}{2N} \left( \sum_{k=1}^{N}\log\frac{\exp(s(i^+_k,t^+_k)/\tau)}{\sum_{t \in \tilde{T}}\exp(s(i^+_k,t)/\tau)} + \sum_{k=1}^{N}\log\frac{\exp(s(i^-_k,t^-_k)/\tau)}{\sum_{t \in \tilde{T}}\exp(s(i^-_k,t)/\tau)} \right) \\
    &= \frac{1}{2} \left( \mathcal{L}_{NCE}^{I^+ \rightarrow T^+; T^-} + \mathcal{L}_{NCE}^{I^- \rightarrow T^-; T^+} \right).
\end{align}

Similarly, for the Text-to-Image direction $\mathcal{L}_{NCE}^{\tilde{T} \rightarrow \tilde{I}}$, we can derive:
\begin{equation}
    \mathcal{L}_{NCE}^{\tilde{T} \rightarrow \tilde{I}} = \frac{1}{2} \left( \mathcal{L}_{NCE}^{T^+ \rightarrow I^+; I^-} + \mathcal{L}_{NCE}^{T^- \rightarrow I^-; I^+} \right).
\end{equation}

Combining these, the total CLIP loss can be expressed as:
\begin{equation}
    \mathcal{L}_{CLIP}(\tilde{I}, \tilde{T}) = \frac{1}{4} \left( \mathcal{L}_{NCE}^{I^+ \rightarrow T^+; T^-} + \mathcal{L}_{NCE}^{I^- \rightarrow T^-; T^+} + \mathcal{L}_{NCE}^{T^+ \rightarrow I^+; I^-} + \mathcal{L}_{NCE}^{T^- \rightarrow I^-; I^+} \right).
\end{equation}

However, the TripletCLIP loss removes the noisy Text-to-Image comparisons involving hard negatives (i.e., $T^+ \rightarrow I^-$ and $T^- \rightarrow I^+$). We can express the relation between the standard CLIP terms and the "clean" Triplet terms with an additional cost $C \geq 0$:

\begin{align}
    \mathcal{L}_{NCE}^{T^+ \rightarrow I^+; I^-} &= \mathcal{L}_{NCE}^{T^+ \rightarrow I^+} + C_{T^+ \rightarrow I^-} \\
    \mathcal{L}_{NCE}^{T^- \rightarrow I^-; I^+} &= \mathcal{L}_{NCE}^{T^- \rightarrow I^-} + C_{T^- \rightarrow I^+}
\end{align}

Where $C$ represents the log-sum-exp term of the additional negatives included in the denominator.
Thus, we conclude that the standard CLIP loss is lower-bounded by the TripletCLIP loss:

\begin{equation}
    \mathcal{L}_{CLIP}(\tilde{I}, \tilde{T}) = \frac{1}{4} \left( \mathcal{L}_{TripletCLIP}(\tilde{I}, \tilde{T}) + C_{T^+ \rightarrow I^-} + C_{T^- \rightarrow I^+} \right).
\end{equation}

\section{Complete PolyGen Results}
\label{app:results}

\begin{table}[t]
    \scriptsize
    \centering
    \begin{tabularx}{\textwidth}{ccccc YYY YY YY Y YYY YY Y}
    \toprule
     & & \multicolumn{3}{c}{\textbf{Settings}} & \multicolumn{8}{c}{\textbf{SugarCrepe}} & \multicolumn{6}{c}{\textbf{SugarCrepe++ (ITT)}} \\
    \cmidrule(lr){3-5} \cmidrule(lr){6-13} \cmidrule(lr){14-19}
     & & & & & \multicolumn{3}{c}{Replace} & \multicolumn{2}{c}{Swap} & \multicolumn{2}{c}{Add} & & \multicolumn{3}{c}{Replace} & \multicolumn{2}{c}{Swap} & \\
    $n^+$ & $m$ & HN & $\mathcal{L}_{HN}$ & S & Att & Obj & Rel & Att & Obj & Att & Obj & \textbf{Avg} & Att & Obj & Rel & Att & Obj & \textbf{Avg} \\
    \midrule
    1 & 1 & {\color{pred}{\xmark}} & {\color{pred}{\xmark}} & {\color{pred}{\xmark}} & 60.0 & 64.2 & 53.2 & 51.2 & 48.2 & 59.0 & 52.0 & 55.4 & 42.1 & 46.7 & 33.2 & 31.5 & 29.8 & 36.7 \\
    \addlinespace
    2 & 2 & {\color{pgreen}{\checkmark}} & {\color{pred}{\xmark}} & {\color{pred}{\xmark}} & 64.6 & 60.6 & 53.7 & 52.9 & \textbf{57.1} & 58.7 & 50.9 & 56.9 & 49.0 & 47.6 & 35.9 & 33.0 & 34.3 & 40.0 \\
    2 & 2 & {\color{pgreen}{\checkmark}} & {\color{pred}{\xmark}} & {\color{pgreen}{\checkmark}} & 64.9 & 61.7 & \textbf{54.6} & 49.1 & 46.9 & 56.5 & 51.5 & 55.0 & 48.9 & 49.2 & 36.7 & 31.2 & 28.2 & 38.8 \\
    2 & 2 & {\color{pgreen}{\checkmark}} & {\color{pgreen}{\checkmark}} & {\color{pred}{\xmark}} & 63.1 & 61.4 & 51.4 & 52.6 & 49.8 & 61.3 & 50.8 & 55.8 & 49.8 & 47.3 & 33.9 & 32.1 & 32.7 & 39.2 \\
    2 & 2 & {\color{pgreen}{\checkmark}} & {\color{pgreen}{\checkmark}} & {\color{pgreen}{\checkmark}} & 64.6 & 61.4 & 52.4 & 49.0 & 53.5 & \textbf{61.9} & 51.8 & 56.4 & \textbf{50.6} & 47.6 & 35.5 & 31.1 & \textbf{37.6} & 40.5 \\
    3 & 3 & {\color{pgreen}{\checkmark}} & {\color{pgreen}{\checkmark}} & {\color{pgreen}{\checkmark}} & \textbf{65.0} & \textbf{64.9} & 53.6 & \textbf{54.7} & \textbf{57.1} & 58.2 & \textbf{52.9} & \textbf{58.1} & 50.3 & \textbf{52.9} & 36.0 & \textbf{36.0} & 32.7 & \textbf{41.6} \\
    4 & 4 & {\color{pgreen}{\checkmark}} & {\color{pgreen}{\checkmark}} & {\color{pgreen}{\checkmark}} & 63.8 & 64.2 & 52.4 & 52.1 & 51.0 & \textbf{61.9} & 51.9 & 56.8 & \textbf{50.6} & 51.7 & \textbf{37.0} & 33.3 & 31.0 & 40.7 \\
    \bottomrule
    \end{tabularx}
    \caption{\textbf{Fine-grained syntactic binding evaluation.} Results on SugarCrepe benchmarks demonstrate that maximizing ensemble heterogeneity ($m$) is crucial for resolving complex compositional ambiguities, preventing the model from relying on lexical shortcuts.}
    \label{tab:sugarcrepe-full}
\end{table}

\begin{table}[t]
    \centering
    \small
    \begin{tabularx}{\textwidth}{cc ccc YYYY YYYY Y}
        \toprule
         & & \multicolumn{3}{c}{\textbf{HN Comp.}} & \multicolumn{4}{c}{\textbf{Image Retrieval}} & \multicolumn{4}{c}{\textbf{Text Retrieval}} & \textbf{0-shot} \\
        \cmidrule(lr){3-5} \cmidrule(lr){6-9} \cmidrule(lr){10-13} \cmidrule(lr){14-14}
        $n^+$ & $m$ & Smpl & Loss & Sched & MS Coco & F8K & F30K & \textbf{Avg} & MS Coco & F8K & F30K & \textbf{Avg} & ImgNet \\
        \midrule
        1 & 1 & {\color{pred}{\xmark}} & {\color{pred}{\xmark}} & {\color{pred}{\xmark}} & 6.4 & 15.0 & 13.3 & 11.6 & 8.7 & 21.5 & 17.5 & 15.9 & 8.15 \\
        \addlinespace
        2 & 2 & {\color{pgreen}{\checkmark}} & {\color{pred}{\xmark}} & {\color{pred}{\xmark}} & 6.7 & 16.7 & 16.1 & 13.2 & 8.9 & 23.3 & 21.5 & 17.9 & 9.95 \\
        2 & 2 & {\color{pgreen}{\checkmark}} & {\color{pred}{\xmark}} & {\color{pgreen}{\checkmark}} & 7.5 & 16.4 & 16.9 & 13.6 & 10.4 & 23.0 & 22.9 & 18.8 & 10.23 \\
        2 & 2 & {\color{pgreen}{\checkmark}} & {\color{pgreen}{\checkmark}} & {\color{pred}{\xmark}} & 7.8 & 16.9 & 17.4 & 14.0 & 9.8 & 22.8 & 24.4 & 19.0 & 10.27 \\
        2 & 2 & {\color{pgreen}{\checkmark}} & {\color{pgreen}{\checkmark}} & {\color{pgreen}{\checkmark}} & 7.7 & 18.3 & 17.8 & 14.6 & 10.5 & \textbf{26.2} & \textbf{24.9} & \textbf{20.5} & 10.80 \\
        3 & 3 & {\color{pgreen}{\checkmark}} & {\color{pgreen}{\checkmark}} & {\color{pgreen}{\checkmark}} & 8.7 & \textbf{19.5} & 18.8 & \textbf{15.7} & 10.1 & 22.9 & 23.6 & 18.9 & 10.72 \\
        4 & 4 & {\color{pgreen}{\checkmark}} & {\color{pgreen}{\checkmark}} & {\color{pgreen}{\checkmark}} & \textbf{8.8} & 18.9 & \textbf{19.4} & \textbf{15.7} & \textbf{11.5} & 24.9 & 24.7 & 20.4 & \textbf{11.01} \\
        \bottomrule
    \end{tabularx}
    \caption{\textbf{Hard Negative strategy analysis.} The combination of Curriculum Scheduler and Triplet Loss optimizes the stability-plasticity trade-off, mitigating early-training gradient noise while imposing a stricter margin for retrieval tasks.}
    \label{tab:retrieval_main_hn}
\end{table}

We provide here the granular breakdown of performance across all downstream tasks. \Cref{tab:sugarcrepe-full} details the compositional reasoning scores on SugarCrepe and SugarCrepe++. \Cref{tab:retrieval_main_hn} analyzes the impact of Hard Negative strategies on cross-modal retrieval. Finally, \Cref{tab:vision-main} and \Cref{tab:retrieval-main} report the full per-dataset metrics for Vision-Only transfer (Linear Probing, Few-Shot) and Zero-Shot Classification/Retrieval, respectively, while \Cref{tab:vision_main_hn} presents the ablation study on vision tasks.

\begin{table}[t]
    \centering
    \footnotesize
    \begin{tabularx}{\textwidth}{llccYYYYYYYYYY}
        \toprule
        & & & & \multicolumn{9}{c}{\textbf{Vision Datasets}} & \\
        \cmidrule(lr){5-13}
        \textbf{Method} & $n^+$ & $m$ & $\mathcal{L}_{I2I}$ & \rotatebox{90}{CIFAR10} & \rotatebox{90}{CIFAR100} & \rotatebox{90}{Aircraft} & \rotatebox{90}{DTD} & \rotatebox{90}{Flowers} & \rotatebox{90}{Pets} & \rotatebox{90}{SUN397} & \rotatebox{90}{Caltech} & \rotatebox{90}{Food} & \textbf{Avg} \\
        \midrule
        \multicolumn{14}{l}{\textit{\textbf{Linear Probing}}} \\
        \addlinespace[2pt]
         & 1 & 1 & {\color{pred}{\xmark}} & 72.5 & 50.8 & 24.7 & \textbf{47.2} & 69.7 & 38.9 & 39.6 & 51.9 & 46.5 & 49.1 \\
         & 1 & 2 & {\color{pred}{\xmark}} & 72.9 & 51.0 & 20.4 & 44.1 & 67.6 & 36.7 & 38.0 & 51.7 & 45.4 & 47.5 \\
         & 1 & 3 & {\color{pred}{\xmark}} & 73.0 & 51.6 & 23.5 & 44.3 & 70.9 & 37.4 & 38.1 & 51.9 & 45.8 & 48.5 \\
         & 1 & 4 & {\color{pred}{\xmark}} & 73.1 & 51.4 & 23.8 & 44.7 & 71.1 & 36.8 & 39.1 & 53.5 & 46.8 & 48.9 \\
        \addlinespace
         & 2 & 1 & {\color{pred}{\xmark}} & 72.2 & 50.3 & 23.4 & 47.1 & \textbf{71.6} & 39.2 & 40.3 & 54.5 & 47.4 & 49.5 \\
         & 2 & 2 & {\color{pred}{\xmark}} & 73.0 & 49.0 & 22.2 & 45.9 & 68.3 & 36.2 & 39.4 & 51.0 & 46.2 & 47.9 \\
         & 3 & 1 & {\color{pred}{\xmark}} & 72.8 & 51.3 & 24.6 & 46.9 & 70.6 & \textbf{39.8} & \textbf{41.1} & 50.9 & 46.1 & 49.3 \\
         & 3 & 3 & {\color{pred}{\xmark}} & 73.6 & 51.6 & 22.8 & 45.2 & 68.5 & 35.4 & 39.0 & 51.1 & 46.8 & 48.2 \\
         & 4 & 1 & {\color{pred}{\xmark}} & 72.8 & 51.3 & \textbf{25.3} & 46.9 & 70.2 & 38.9 & 40.9 & \textbf{54.9} & 47.3 & \textbf{49.8} \\
         & 4 & 4 & {\color{pred}{\xmark}} & \textbf{73.9} & \textbf{51.8} & 23.9 & 45.1 & 69.0 & 37.7 & 40.1 & 50.5 & \textbf{47.6} & 48.9 \\
        \addlinespace
         & 2 & 1 & {\color{pgreen}{\checkmark}} & 70.5 & 47.2 & 20.7 & 41.5 & 57.7 & 31.9 & 34.4 & 42.8 & 43.6 & 43.4 \\
         & 2 & 2 & {\color{pgreen}{\checkmark}} & 69.8 & 47.6 & 19.4 & 37.3 & 54.6 & 31.0 & 32.6 & 40.2 & 40.7 & 41.5 \\
         & 3 & 1 & {\color{pgreen}{\checkmark}} & 71.7 & 49.2 & 20.9 & 43.5 & 58.4 & 34.9 & 36.0 & 42.8 & 42.8 & 44.4 \\
         & 3 & 3 & {\color{pgreen}{\checkmark}} & 70.5 & 48.3 & 19.5 & 37.6 & 52.8 & 31.5 & 31.8 & 39.1 & 40.5 & 41.3 \\
         & 4 & 1 & {\color{pgreen}{\checkmark}} & 72.4 & 50.7 & 21.4 & 44.5 & 57.8 & 34.9 & 36.2 & 43.7 & 45.0 & 45.2 \\
         & 4 & 4 & {\color{pgreen}{\checkmark}} & 71.6 & 49.7 & 19.9 & 41.1 & 56.0 & 31.5 & 33.2 & 39.8 & 43.0 & 42.9 \\
        \midrule
        \multicolumn{14}{l}{\textit{\textbf{Few-shot (5-shot, 5-way)}}} \\
        \addlinespace[2pt]
         & 1 & 1 & {\color{pred}{\xmark}} & 46.3 & 57.8 & \textbf{39.8} & 63.5 & 85.6 & 48.5 & 84.2 & 75.4 & 53.4 & 61.6 \\
         & 1 & 2 & {\color{pred}{\xmark}} & 47.8 & 59.5 & 39.1 & 60.9 & 85.1 & 47.0 & 83.8 & 75.3 & 51.2 & 61.1 \\
         & 1 & 3 & {\color{pred}{\xmark}} & 48.2 & 59.2 & 39.1 & 61.7 & 85.8 & 47.3 & 83.8 & 75.4 & 51.5 & 61.3 \\
         & 1 & 4 & {\color{pred}{\xmark}} & 47.7 & \textbf{59.7} & 39.3 & 61.7 & 85.9 & 47.5 & 84.3 & 75.4 & 52.4 & 61.5 \\
        \addlinespace
         & 2 & 1 & {\color{pred}{\xmark}} & 46.2 & 58.1 & 38.8 & 65.1 & \textbf{87.0} & 48.9 & 86.9 & 77.4 & 53.4 & 62.4 \\
         & 2 & 2 & {\color{pred}{\xmark}} & 48.0 & 59.3 & 37.7 & 61.4 & 84.9 & 47.1 & 85.1 & 74.6 & 52.1 & 61.1 \\
         & 3 & 1 & {\color{pred}{\xmark}} & 47.0 & 58.5 & 38.9 & 65.4 & 86.0 & \textbf{49.2} & 87.4 & 78.1 & 52.1 & 62.7 \\
         & 3 & 3 & {\color{pred}{\xmark}} & 47.9 & 59.1 & 37.4 & 62.5 & 84.1 & 46.3 & 85.7 & 76.0 & 51.3 & 61.1 \\
         & 4 & 1 & {\color{pred}{\xmark}} & 47.0 & 58.3 & 38.6 & \textbf{65.7} & 86.0 & 48.7 & \textbf{88.0} & \textbf{78.7} & \textbf{54.5} & \textbf{62.8} \\
         & 4 & 4 & {\color{pred}{\xmark}} & \textbf{49.5} & \textbf{59.7} & 38.1 & 62.9 & 84.3 & 47.7 & 86.7 & 77.0 & 52.2 & 62.0 \\
        \addlinespace
         & 2 & 1 & {\color{pgreen}{\checkmark}} & 43.9 & 53.9 & 36.3 & 57.7 & 78.3 & 42.7 & 84.5 & 71.6 & 47.1 & 57.3 \\
         & 2 & 2 & {\color{pgreen}{\checkmark}} & 43.0 & 54.0 & 35.5 & 54.8 & 75.9 & 40.6 & 82.8 & 69.6 & 45.1 & 55.7 \\
         & 3 & 1 & {\color{pgreen}{\checkmark}} & 42.9 & 53.0 & 36.0 & 57.8 & 78.1 & 42.7 & 85.3 & 72.1 & 48.2 & 57.3 \\
         & 3 & 3 & {\color{pgreen}{\checkmark}} & 43.1 & 52.6 & 34.8 & 54.2 & 76.0 & 39.1 & 82.4 & 69.5 & 45.1 & 55.2 \\
         & 4 & 1 & {\color{pgreen}{\checkmark}} & 43.2 & 54.2 & 34.5 & 58.6 & 78.3 & 42.5 & 86.0 & 73.2 & 48.8 & 57.7 \\
         & 4 & 4 & {\color{pgreen}{\checkmark}} & 44.8 & 53.9 & 34.8 & 55.3 & 77.5 & 40.3 & 84.3 & 70.0 & 46.8 & 56.4 \\
        \bottomrule
    \end{tabularx}
    \caption{\textbf{Vision-Only representational quality.} Linear probing gains correspond strongly to positive sample density ($n^+$), indicating that stabilizing the visual cluster centroid is more effective for linear separability than introducing hard negative discriminators.}
    \label{tab:vision-main}
\end{table}

\begin{table}[t]
    \centering
    \small
    \begin{tabularx}{\textwidth}{ccc YYYY YYYY c}
        \toprule
         & & & \multicolumn{4}{c}{\textbf{Image Retrieval}} & \multicolumn{4}{c}{\textbf{Text Retrieval}} & \textbf{0-shot} \\
         \cmidrule(lr){4-7} \cmidrule(lr){8-11} \cmidrule(lr){12-12}
        $n^+$ & $m$ & $\mathcal{L}_{I2I}$ & MS Coco & Flkr 8K & Flkr 30K & \textbf{Avg} & MS Coco & Flkr 8K & Flkr 30K & \textbf{Avg} & ImgNet \\
        \midrule
        1 & 1 & {\color{pred}{\xmark}} & 6.4 & 15.0 & 13.3 & 11.6 & 8.7 & 21.5 & 17.5 & 15.9 & 8.15 \\
        1 & 2 & {\color{pred}{\xmark}} & 7.1 & 15.6 & 13.8 & 12.1 & 9.8 & 22.1 & 20.6 & 17.5 & 8.53 \\
        1 & 3 & {\color{pred}{\xmark}} & 7.4 & 17.1 & 15.3 & 13.3 & 9.4 & 24.1 & 20.2 & 17.9 & 8.56 \\
        1 & 4 & {\color{pred}{\xmark}} & 7.2 & 17.0 & 14.3 & 12.8 & 9.4 & 23.3 & 20.2 & 17.6 & 8.92 \\
        \addlinespace
        2 & 1 & {\color{pred}{\xmark}} & 7.5 & 17.2 & 16.8 & 13.8 & 9.5 & 23.8 & 22.9 & 18.7 & 9.41 \\
        2 & 2 & {\color{pred}{\xmark}} & 7.6 & 17.7 & 17.2 & 14.2 & 10.4 & 25.8 & 24.3 & 20.2 & 9.54 \\
        3 & 1 & {\color{pred}{\xmark}} & 8.9 & 17.6 & 17.5 & 14.7 & \textbf{12.1} & 26.6 & 24.7 & 21.1 & 9.66 \\
        3 & 3 & {\color{pred}{\xmark}} & 8.9 & \textbf{20.6} & 19.1 & 16.2 & \textbf{12.1} & \textbf{29.7} & 25.4 & \textbf{22.4} & 9.38 \\
        4 & 1 & {\color{pred}{\xmark}} & 8.1 & 17.9 & 18.4 & 14.8 & 10.5 & 23.0 & 24.7 & 19.4 & 9.56 \\
        4 & 4 & {\color{pred}{\xmark}} & \textbf{9.0} & 19.9 & 19.4 & 16.1 & 11.4 & 27.1 & 25.8 & 21.4 & 9.86 \\
        \addlinespace
        2 & 1 & {\color{pgreen}{\checkmark}} & 7.3 & 16.2 & 16.4 & 13.3 & 9.1 & 21.6 & 22.2 & 17.6 & 10.14 \\
        2 & 2 & {\color{pgreen}{\checkmark}} & 8.2 & 17.4 & 17.7 & 14.4 & 11.1 & 24.5 & 23.7 & 19.8 & 11.19 \\
        3 & 1 & {\color{pgreen}{\checkmark}} & 7.9 & 17.5 & 17.3 & 14.2 & 10.0 & 22.0 & 24.7 & 18.9 & 10.88 \\
        3 & 3 & {\color{pgreen}{\checkmark}} & \textbf{9.0} & 20.2 & \textbf{20.3} & \textbf{16.5} & 11.5 & 26.4 & \textbf{27.0} & 21.7 & 10.98 \\
        4 & 1 & {\color{pgreen}{\checkmark}} & 8.0 & 18.6 & 18.3 & 15.0 & 10.4 & 24.4 & 24.8 & 19.9 & 11.24 \\
        4 & 4 & {\color{pgreen}{\checkmark}} & 8.8 & 18.9 & 19.4 & 15.7 & 11.1 & 24.5 & 26.2 & 20.6 & \textbf{11.34} \\
        \bottomrule
    \end{tabularx}
    \caption{\textbf{Zero-Shot and Retrieval performance.} Vision-Language generalization correlates with ensemble size ($m$) and invariance learning ($\mathcal{L}_{I2I}$), confirming that marginalizing generator-specific artifacts improves alignment with the open-world semantic manifold.}
    \label{tab:retrieval-main}
\end{table}

\begin{table}[t]
    \centering
    \footnotesize
    \begin{tabularx}{\textwidth}{cc ccc YYYYYYYYYY}
        \toprule
         & & \multicolumn{3}{c}{\textbf{HN Components}} & \multicolumn{9}{c}{\textbf{Vision Datasets}} & \\
         \cmidrule(lr){3-5} \cmidrule(lr){6-14}
        $n^+$ & $m$ & Smpl & Loss & Sched & C10 & C100 & Air & DTD & Flwr & Pets & SUN & Cal & Food & \textbf{Avg} \\
        \midrule
        \multicolumn{15}{l}{\textit{\textbf{Linear Probing}}} \\
        1 & 1 & {\color{pred}{\xmark}} & {\color{pred}{\xmark}} & {\color{pred}{\xmark}} & \textbf{72.5} & \textbf{50.8} & \textbf{24.7} & \textbf{47.2} & \textbf{69.7} & \textbf{38.9} & \textbf{39.6} & \textbf{51.9} & \textbf{46.5} & \textbf{49.1} \\
        \addlinespace
        2 & 2 & {\color{pgreen}{\checkmark}} & {\color{pred}{\xmark}} & {\color{pred}{\xmark}} & 70.4 & 46.9 & 19.8 & 38.4 & 53.4 & 30.6 & 31.9 & 41.6 & 40.4 & 41.5 \\
        2 & 2 & {\color{pgreen}{\checkmark}} & {\color{pred}{\xmark}} & {\color{pgreen}{\checkmark}} & 69.6 & 45.7 & 19.2 & 37.7 & 54.4 & 29.8 & 31.9 & 39.4 & 39.3 & 40.8 \\
        2 & 2 & {\color{pgreen}{\checkmark}} & {\color{pgreen}{\checkmark}} & {\color{pred}{\xmark}} & 71.0 & 48.9 & 18.9 & 38.7 & 55.8 & 31.4 & 33.2 & 42.1 & 41.2 & 42.3 \\
        2 & 2 & {\color{pgreen}{\checkmark}} & {\color{pgreen}{\checkmark}} & {\color{pgreen}{\checkmark}} & 70.2 & 47.4 & 20.0 & 38.0 & 54.6 & 29.0 & 32.3 & 41.1 & 39.8 & 41.4 \\
        3 & 3 & {\color{pgreen}{\checkmark}} & {\color{pgreen}{\checkmark}} & {\color{pgreen}{\checkmark}} & 71.1 & 48.6 & 20.0 & 34.2 & 53.4 & 29.0 & 31.7 & 38.1 & 41.6 & 40.8 \\
        4 & 4 & {\color{pgreen}{\checkmark}} & {\color{pgreen}{\checkmark}} & {\color{pgreen}{\checkmark}} & 71.8 & 49.9 & 19.5 & 39.5 & 55.8 & 31.8 & 33.4 & 39.8 & 43.2 & 42.7 \\
        \midrule
        \multicolumn{15}{l}{\textit{\textbf{Few-shot}}} \\
        1 & 1 & {\color{pred}{\xmark}} & {\color{pred}{\xmark}} & {\color{pred}{\xmark}} & \textbf{46.3} & \textbf{57.8} & \textbf{39.8} & \textbf{63.5} & \textbf{85.6} & \textbf{48.5} & \textbf{84.2} & \textbf{75.4} & \textbf{53.4} & \textbf{61.6} \\
        \addlinespace
        2 & 2 & {\color{pgreen}{\checkmark}} & {\color{pred}{\xmark}} & {\color{pred}{\xmark}} & 42.4 & 53.9 & 34.8 & 54.6 & 76.8 & 41.5 & 82.7 & 70.0 & 44.9 & 55.7 \\
        2 & 2 & {\color{pgreen}{\checkmark}} & {\color{pred}{\xmark}} & {\color{pgreen}{\checkmark}} & 43.2 & 54.5 & 34.1 & 54.3 & 76.7 & 41.6 & 83.0 & 69.7 & 45.1 & 55.8 \\
        2 & 2 & {\color{pgreen}{\checkmark}} & {\color{pgreen}{\checkmark}} & {\color{pred}{\xmark}} & 43.4 & 55.3 & 34.9 & 55.5 & 78.0 & 42.5 & 84.0 & 70.7 & 46.4 & 56.8 \\
        2 & 2 & {\color{pgreen}{\checkmark}} & {\color{pgreen}{\checkmark}} & {\color{pgreen}{\checkmark}} & 42.2 & 54.0 & 34.8 & 54.5 & 77.0 & 41.2 & 83.1 & 69.7 & 45.5 & 55.8 \\
        3 & 3 & {\color{pgreen}{\checkmark}} & {\color{pgreen}{\checkmark}} & {\color{pgreen}{\checkmark}} & 44.3 & 54.5 & 34.5 & 53.9 & 76.4 & 39.0 & 82.3 & 69.3 & 45.3 & 55.5 \\
        4 & 4 & {\color{pgreen}{\checkmark}} & {\color{pgreen}{\checkmark}} & {\color{pgreen}{\checkmark}} & 44.8 & 55.2 & 35.8 & 54.7 & 77.8 & 40.1 & 83.9 & 70.0 & 46.4 & 56.5 \\
        \bottomrule
    \end{tabularx}
    \caption{\textbf{Ablation of Hard Negative integration.} The results isolate the Curriculum Scheduler as the critical component for preventing feature degradation in standard vision tasks, shielding the representation from noisy negative gradients during early alignment.}
    \label{tab:vision_main_hn}
\end{table}

\section{Generative Models Selection}
\label{app:selection}

\subsection{LLM selection for hard negative captions - attribute classification}
\label{app:llm_hn_selection}

\begin{figure}[t]
    \centering
    \includegraphics[width=\textwidth]{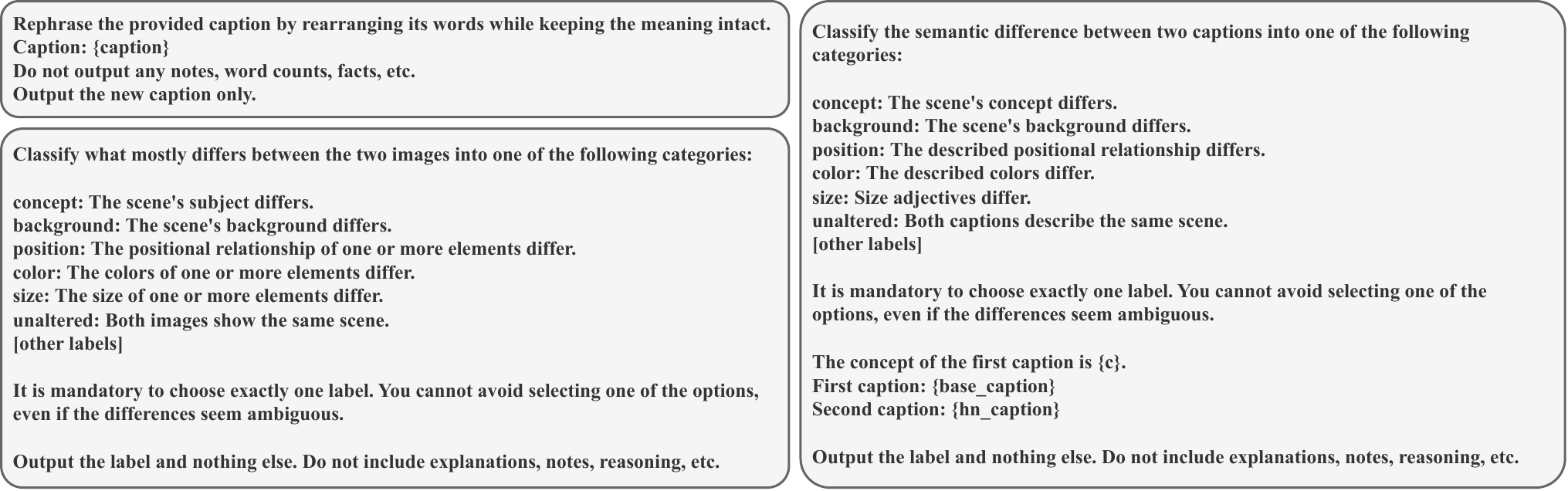}
    \caption{\textbf{Verification protocols for synthetic data.} Zero-shot classification prompts used to quantitatively assess instruction-following adherence and rule out hallucinations in the generation pipeline.}
    \label{fig:class_prompts}
\end{figure}

\begin{figure}[t]
    \centering
    \begin{minipage}{0.48\textwidth}
        \centering
        \resizebox{\textwidth}{!}{
            \begin{tabular}{l | c | c | c | c}
                \hline
                \textbf{Model} & \textbf{Accuracy} & \textbf{Avg. F1} & \textbf{Avg. Prec.} & \textbf{Avg. Rec.} \\
                \hline
                Mistral V0.2 & 61\% & 60\% & 76\% & 61\%  \\
                \hline
                Mistral V0.3 & 62\% & 62\% & 77\% & 62\% \\
                \hline
                Qwen 2.5 & 67\% & 67\% & 78\% & 67\%\\
                \hline
                Falcon 7B & 63\% & 65\% & 79\% & 63\% \\
                \hline
                \textbf{Llama 3.1} & \textbf{74\%} & \textbf{75\%} & \textbf{80\%} & \textbf{74\%} \\
                \hline
            \end{tabular}
        }
        \captionof{table}{\textbf{LLM Instruction-Following benchmark.} Llama 3.1 exhibits superior adherence to semantic constraints, minimizing the risk of false negatives (logical contradictions) in the counterfactual caption generation process.}
        \label{tab:hn_class_results}
    \end{minipage}
    \hfill
    \begin{minipage}{0.48\textwidth}
        \centering
        \includegraphics[width=\textwidth]{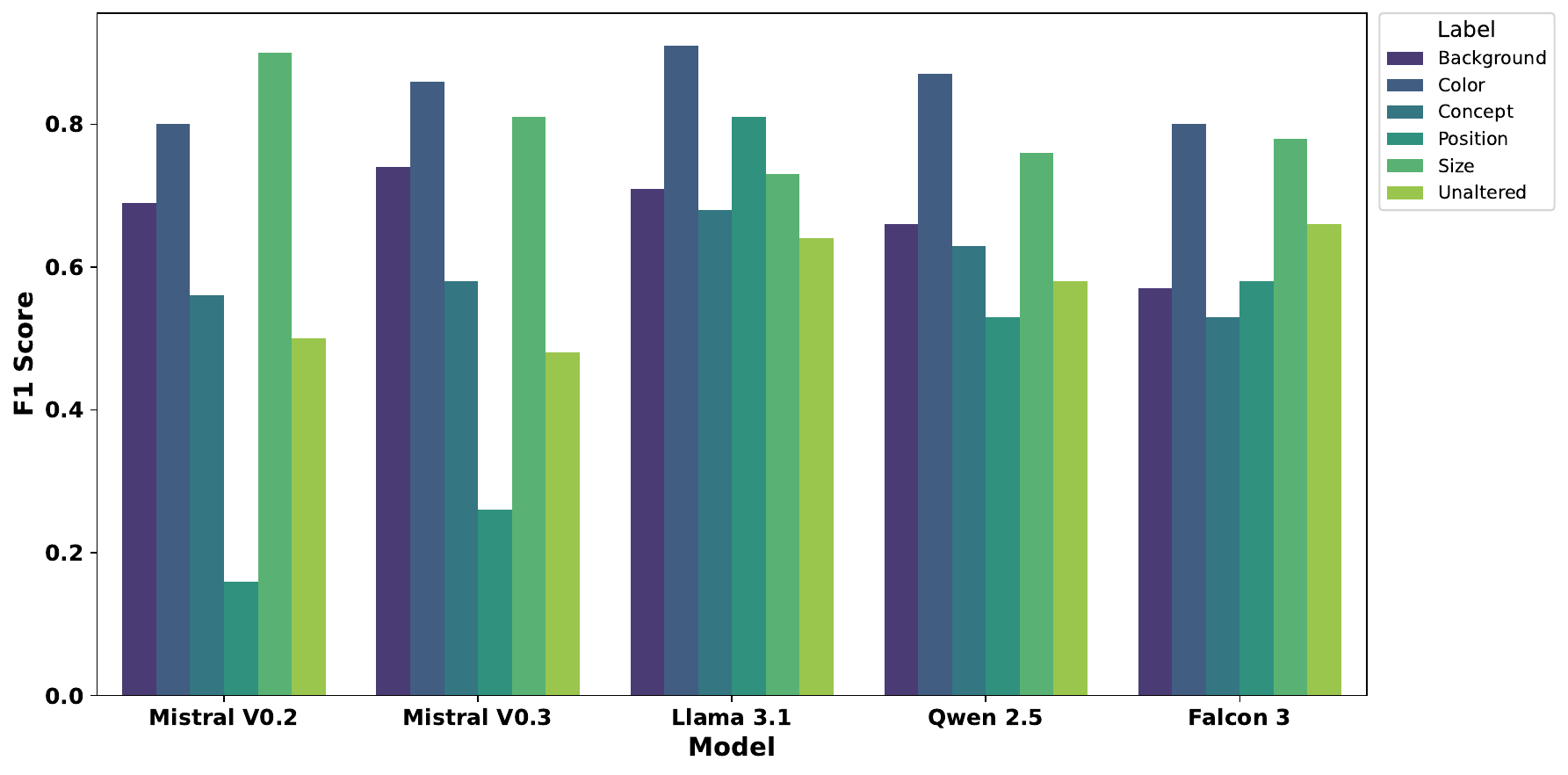}
        \captionof{figure}{\textbf{F1-scores per semantic axis (LLM).} Classification performance across modification types. Some axes (e.g., texture) prove inherently more difficult for LLMs to modify without altering the global context.}
        \label{fig:per_class_f1}
    \end{minipage}
\end{figure}

The task of generating a hard negative caption requires strict instruction following. To select the best LLM for the task, we evaluate a pool $M$ of similar-sized LLMs, namely Mistral-V0.2-7B and Mistral-V0.3-7B, Llama-3.1-8B, Falcon-3-7B, and Qwen-2.5-7B, on a hard-negative attribute classification ensemble. Given the LLM $m\in M$, the concept $c$, the base caption $c_1$, and the hard negative caption $c_2$ generated by $m$ by modifying attribute $a \in A$, we use the prompt defined in \cref{fig:class_prompts} to ask each LLM to classify which attribute of $A$ has been used to generate $c_2$. The final classification is given by majority voting of the predictions made by each model. For each $m \in M$, we generate $4.8k$ pairs of base and hard negative captions. Specifically, we generate 960 base captions for each of a limited set of axes $\mathcal{A}=[\text{\textit{background}}, \text{\textit{\textit{color}}}, \text{\textit{position}}, \text{\textit{size}}, \text{\textit{concept}}]$. Out of those 960 captions, 160 are used for generating an \textit{unaltered} caption, where the model is asked to just rephrase the original caption (\cref{fig:class_prompts}). This is used as a control label to assess the model’s understanding of whether meaningful information has been changes. The remaining 800 captions are used to generate hard negatives. \cref{tab:hn_class_results} shows how Llama greatly overperforms other LLMs in this task, generating captions clearly and strictly modifying the requested axis.

\subsection{LLM selection for baseline captions}
\label{app:llm_selection}
The choice of Mistral-V0.2-7B for generating the base caption set has two different reasons. First, for a fairer comparison to the SynthCLIP pipeline, which uses the same model to generate its caption set, shifting the focus towards the prompting strategy and our specific contributions. Secondly, we conduct an ablation on the LLM, generating a $500k$ image-caption pair dataset for both Mistral-V0.2-7B and Llama-3.1-8B, with corresponding images generated by SDXLT. We report a slight advantage on ImageNet zero-shot accuracy with captions generated by Mistral ($1.8\%$ top-1, $5.97\%$ top-5) over Llama ($1.66\%$ top-1, $5.61\%$ top-5), further validating our choice.

\subsection{Diffusion models selection}
\label{app:t2i_selection}
\begin{figure}[t]
    \centering
    \begin{minipage}{0.48\textwidth}
        \centering
        \resizebox{\textwidth}{!}{
            \begin{tabular}{l| c | c | c | c | c}
            \hline
            \textbf{Model} & \textbf{Res.} & \textbf{Steps} & \textbf{FID} $\downarrow$ & \textbf{CLIPScore} $\uparrow$ & \textbf{TIFA} $\uparrow$  \\
            \hline \hline
            \textbf{Real} & $512^2$ & - & - & 30.319 & 88.8 \\
            \hline
            \rowcolor{diversity}\textbf{SD 1.5} & $512^2$ & 50 & 23.014 & 29.962 & 78.7 \\
            \hline
            \rowcolor{recognizability}\textbf{SDXLT} & $512^2$ & 2 & 32.593  & \textbf{31.748} & 87.9 \\
            \hline
            \textbf{SD 2} & $512^2$ & 20 & 53.091 & 27.310 & 69.0 \\
             &  & 50 & 55.348 & 27.244 & 69.0\\
             \cdashline{2-6}
             & $768^2$ & 20 & 23.613 & 30.065 & 80.6 \\
             \rowcolor{diversity} &  & 50 & \textbf{21.095} & 30.162 & 80.4 \\
            \hline
            \textbf{SD 2.1} & $512^2$ & 20 & 61.443 & 27.035 & 68.5 \\
             &  & 50 & 60.213 & 27.128 & 68.8 \\
             \cdashline{2-6}
             & $768^2$ & 20 & 23.047 & 29.991 & 81.3 \\
             \rowcolor{diversity} & & 50 & 23.481 & 30.068 & 82.5 \\
            \hline
            \textbf{SD 3.5 LT} & $512^2$ & 4 & 35.448 & 31.586 & 88.1 \\
             & & 8 & 36.008 & 31.109 & 87.7 \\
             \cdashline{2-6}
             \rowcolor{recognizability}& $1024^2$ & 4 & 36.732 & 31.484 &  89.4 \\
             & & 8 & 36.337 & 31.358 & 89.1 \\
            \hline
            \rowcolor{recognizability}\textbf{SANA} & $512^2$ & 20 & 27.316 & 31.163 & 88.1 \\
            \cdashline{2-6}
             & $1024^2$ & 20 & 27.573 & 31.171 & 88.0 \\
            \hline
            \textbf{FLUX} & $512^2$ & 4 & 31.583 & 31.634 & 90.5\\
            \cdashline{2-6}
            \rowcolor{recognizability} & $1024^2$ & 4 & 28.581 & 31.403 & \textbf{90.9} \\
            \hline
            \end{tabular}
        }
        \caption{\textbf{Quantitative profiling of the generative manifold.} Models are clustered into \textit{Diversity Experts} (high variance) and \textit{Recognizability Experts} (high prompt adherence) to construct a heterogeneous ensemble that covers the visual-semantic pareto frontier.}
        \label{table:t2i_metric_comparison}
    \end{minipage}
    \hfill
    \begin{minipage}{0.48\textwidth}
        \centering
        \resizebox{\textwidth}{!}{
        \begin{tabular}{l | c | c | c | c | c | c}
            \hline
            \textbf{Model} & \textbf{Res.} & \textbf{Steps} & \textbf{Accuracy} & \textbf{Avg F1} & \textbf{Avg Prec.} & \textbf{Avg Rec.} \\
            \hline \hline
            \textbf{SD 1.5} & $512^2$ & 50 & 25\% & 23\% & 28\% & 25\%  \\
            \hline
            \textbf{SDXLT} & $512^2$ & 2 & 37\% & 33\% & 40\% & 37\% \\
            \hline
            \textbf{SD 3.5 LT} & $512^2$ & 4 & 34\% & 30\% & 36\% & 36\% \\
             & & 8 & 32\% & 33\% & 32\% & 32\% \\
             & $1024^2$ & 4 & 35\% & 31\% & 35\% & 35\% \\
             & & 8 & 35\% & 30\% & 36\% & 35\% \\
            \hline
            \textbf{FLUX} & $512^2$ & 4 & 36\% & 33\% & 42\% & 36\% \\
             & $1024^2$ & 4 & \textbf{38\%} & \textbf{35\%} & \textbf{43\%} & \textbf{39\%} \\
            \hline
        \end{tabular}
        }
        \captionof{table}{\textbf{Visual grounding of semantic perturbations.} High-fidelity architectures (Recognizability Experts) show significantly stronger attribute binding capabilities, essential for correctly rendering the fine-grained distinctions required by the Hard Negative curriculum.}
        \label{table:hn_image_classification}
        
        \includegraphics[width=\textwidth]{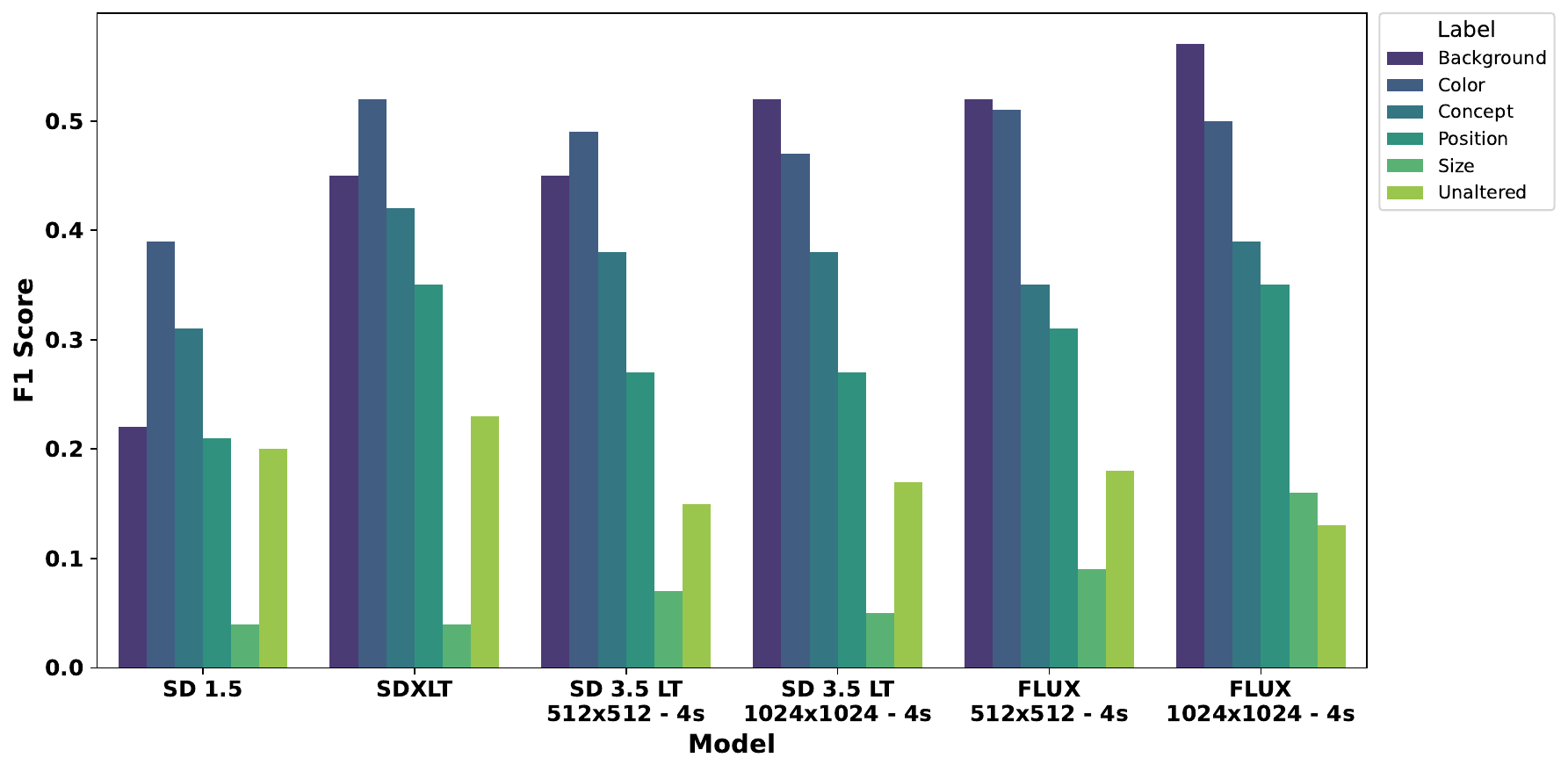}
        \captionof{figure}{\textbf{Evaluation metrics per semantic axis (T2I).} Visual grounding performance across modification types. SDXL-Turbo and FLUX demonstrate superior ability in rendering spatial and object-level changes compared to older architectures.}
        \label{fig:per_class_hn_image_classification}
    \end{minipage}
\end{figure}

\textbf{Image quality metrics.} The selection of the image generators used in PolyGen (SD 1.5, SD 2, SANA, SDXLT) is the result of a comparison of a broader pool of models, also comprising Stable Diffusion v2.1 (SD 2.1), Stable Diffusion 3.5 Large Turbo (SD 3.5 LT), and FLUX.1 Schnell (FLUX).
We randomly sample one caption for each of the $5k$ samples of MS COCO 2017 validation set, and generate the corresponding image with each model. We use these image sets to compare models based on FID, CLIPScore, and TIFA score. Although FID was originally presented as a metric to measure the fidelity of images from GAN models trained on narrow domains, this measure correlates with diversity, as it measures the distribution shift of a set of image features w.r.t corresponding real data. In many diffusion models employing CFG, a low CFG scale is often correlated with lower (better) FID, but also with lower (worse) alignment with the input caption, measured by the CLIPScore, and consequently with lower perceptual quality. For these reasons, ``Diversity expert'' models are related to a low FID, while ``Recognizability experts'' are related to a higher CLIPScore and TIFA. \cref{table:t2i_metric_comparison} shows the results of these evaluations on various resolutions and denoising steps. Although we select the final ensemble taking into account computational constraints, models like FLUX and SD 3.5 could be considered to be used as ``Recognizability experts'' and SD 2.1 as ``Diversity expert''.

\textbf{Semantic axis classification.}
\label{app:t2i_hn_class}
We also test the diffusion models on their ability to correctly and clearly report the variation contained in hard negative captions. To do so, we classify the semantic axis similarly to what we did with the LLMs. This time, we use Qwen 2.5 VL as the classifier, taking as input the image generated from the base caption, the image generated from the hard negative caption, and the image classification prompt shown in \cref{fig:class_prompts}. This time, we use a total of 2.4k pairs of images, 480 for each of the 5 same axes used for testing the LLMs, 80 of which are used for the unaltered class. \cref{table:hn_image_classification} shows how the task of generating useful hard negative images is easier for models with higher image-caption alignment and perceptual quality. Overall, the lower scores compared to the similar experiment on LLMs, due to the noisier signal given by images, justify the use and the performance improvement of the TripletCLIP loss in contrastive models using hard negative samples.

\section{Semantic Axes Selection}
\label{app:axes_selection}

\begin{figure}[t]
    \centering
    \setlength{\tabcolsep}{0pt}
    \subfloat[Llama results. Texture attribute is removed from the final set of semantic axes.\label{fig:cm_llama}]{
        \includegraphics[width=0.475\textwidth,trim=0 0 5 0,clip]{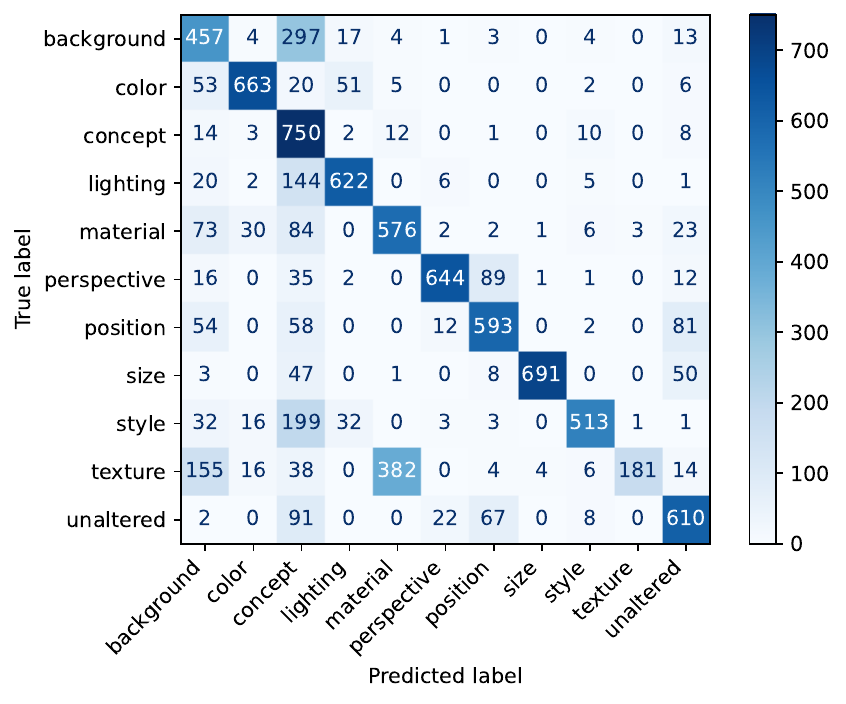}
    }
    \hspace{10pt}
    \subfloat[SDXLT results. Size is removed from the final set of semantic axes.\label{fig:cm_sdxlt}]{
        \includegraphics[width=0.475\textwidth,trim=40 0 25 0,clip]{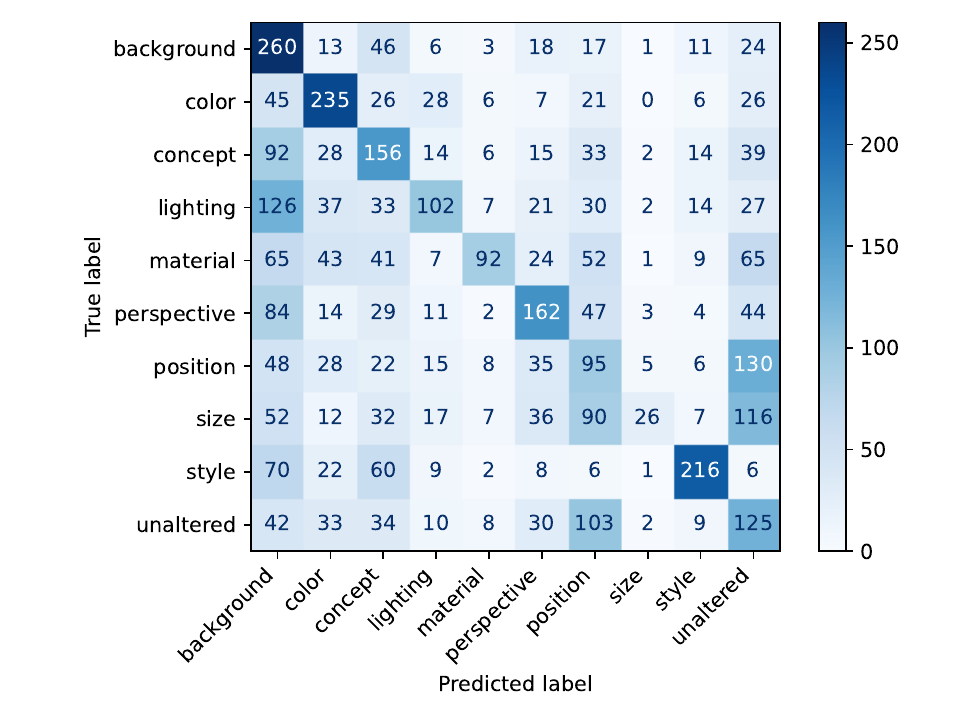}
    }
    \caption{\textbf{Confusion matrices for semantic axis verification.} This analysis drove the pruning of the modification space $\mathcal{A}$, eliminating axes with high confusion rates (e.g., ``Texture'', ``Size'') to ensure the training signal remains structurally consistent.}
    \label{fig:cm}
\end{figure}

Given the results for the semantic axis classification, both to evaluate LLMs and T2I models, we notice how the ability of a model to generate effective hard negatives varies significantly depending on the specific axis being modified (\cref{fig:per_class_f1}, \ref{fig:per_class_hn_image_classification}). The final set of axes is then obtained by a second round of classification from a broader set of axes $A=[\text{\textit{background}}, \text{\textit{color}}, \text{\textit{lighting}}, \text{\textit{material}}, \text{\textit{perspective}}, \text{\textit{position}}, \text{\textit{size}}, \text{\textit{style}}, \text{\textit{texture}}, \text{\textit{concept}}]$, both on pairs of captions generated with Llama, and pairs of images generated by SDXLT. \cref{fig:cm} shows the results of these experiments. We can notice how Llama struggles with generating hard negatives when \textit{texture} is the attribute to be changed. At the same time, SDXLT-generated images struggle to represent the \textit{size} differences between captions. For these reasons, these two attributes are removed from the final set of semantic axes, reported in \cref{fig:prompts}.

\section{T2I Models Comparison in Frequency Domain}
\label{app:spectral_analysis}

\begin{figure}[t]
    \centering
    \subfloat[Radial profiles of the employed models.\label{fig:radial_profiles}]{
        \includegraphics[width=0.48\textwidth]{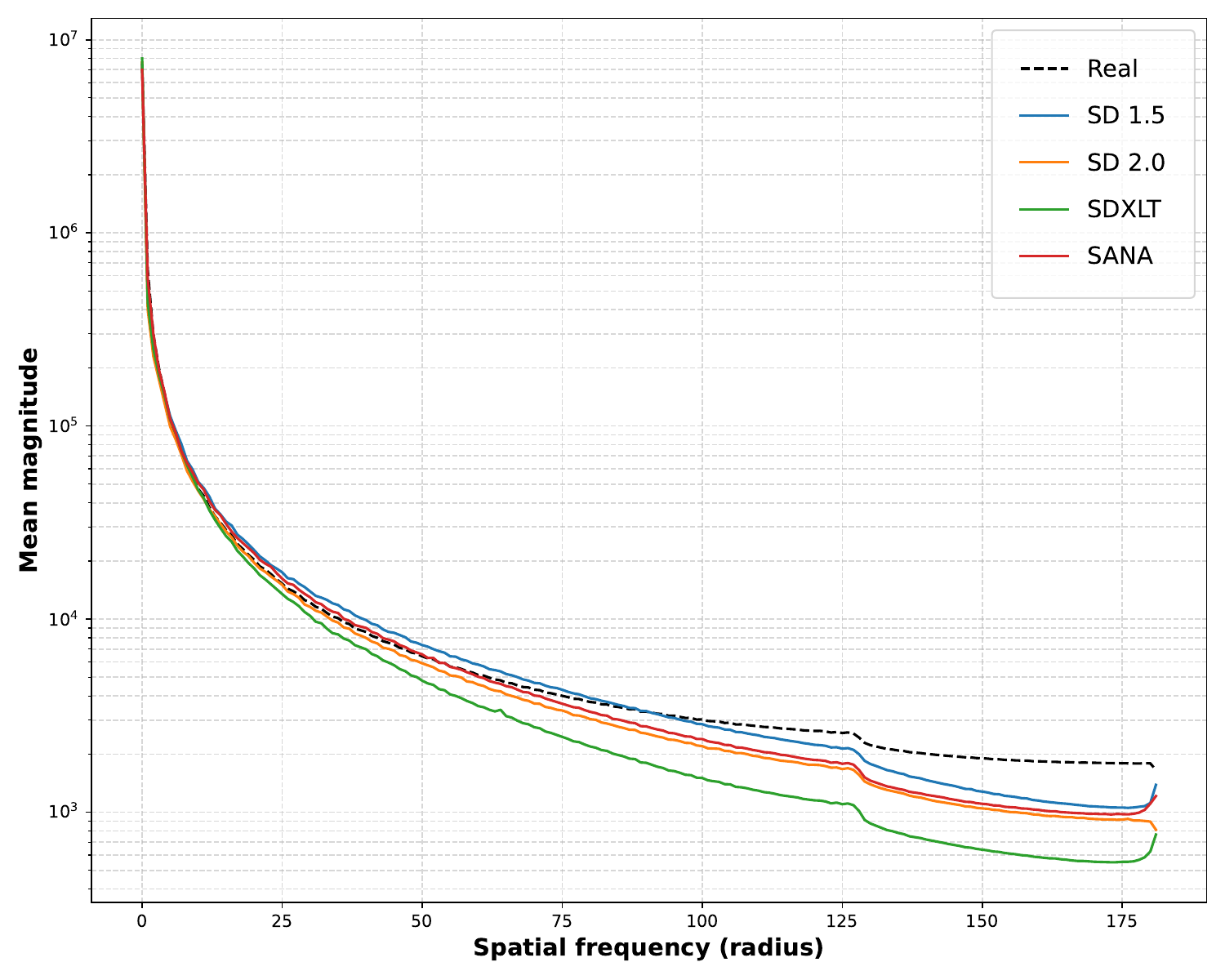}
    }
    \hfill
    \subfloat[Frequency energy distribution of the 50\% highest frequencies.\label{fig:violin_plot}]{
        \includegraphics[width=0.48\textwidth]{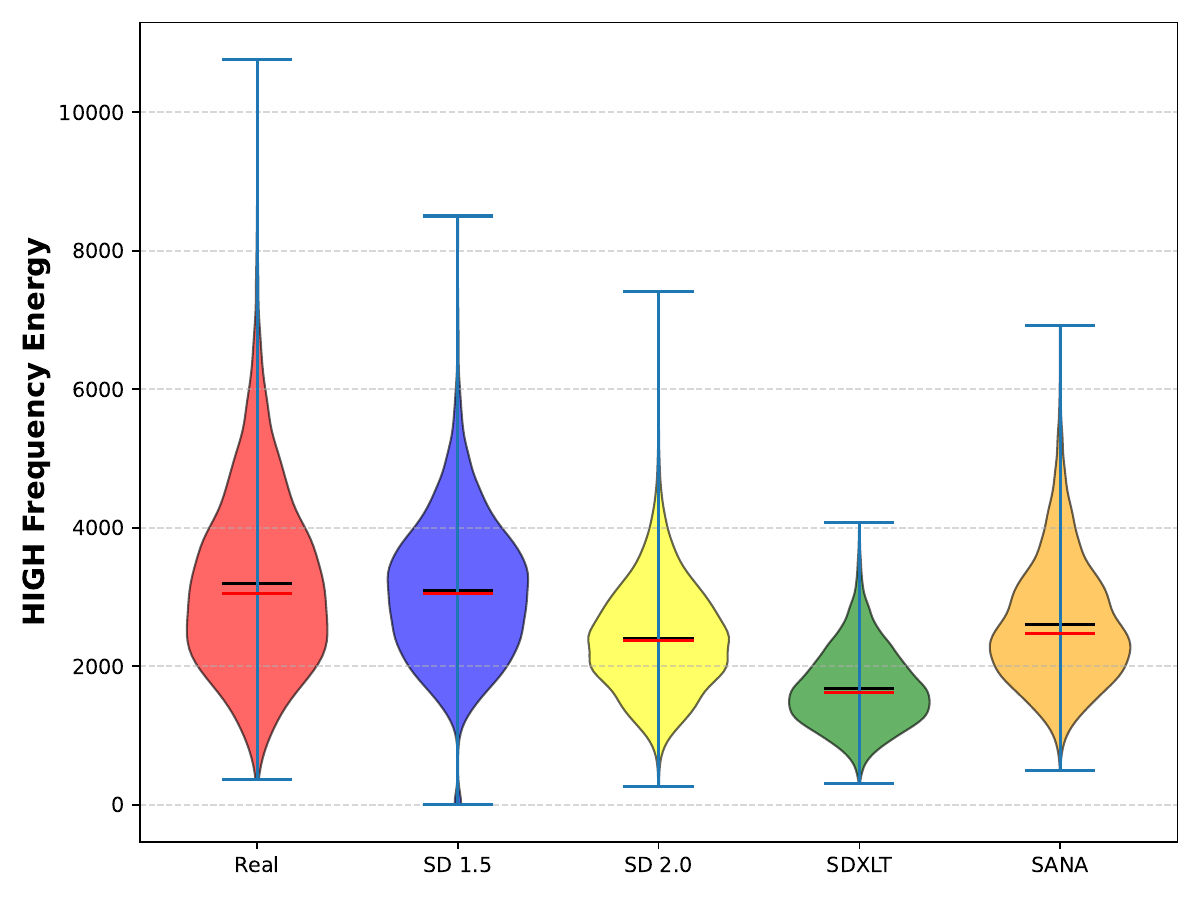}
    }
    \caption{\textbf{Spectral density estimation.} The analysis reveals distinct generator-specific frequency biases (e.g., SDXL high-frequency roll-off). PolyGen trains on the union of these spectra to prevent the model from overfitting to low-level synthesis artifacts.}
    \label{fig:frequency_analysis}
\end{figure}

To get additional insights into the characteristics of the generated images and to try to understand why different diffusion models lead to different training performance, we analyze the images in the frequency domain. This analysis compares the spectral characteristics of images generated by each model against the baseline dataset of real images. To do so, the 2D Fast Fourier Transform (FFT) is applied to each standardized image to convert it from the spatial domain to the frequency domain. From the 2D magnitude spectrum of each image, we extract two key sets of metrics:

\begin{itemize}
    \item \textbf{Radial Frequency Profile:} The 2D spectrum is reduced to a 1D vector using a radial averaging function. This represents the average magnitude as a function of spatial frequency (i.e., the radius from the center). This process is repeated for all $5k$ images in the set, and the mean radial profile for that set is computed.
    \item \textbf{Frequency Energy:} The energy distribution is quantified for each individual image. We calculate the total energy on the high-frequency components $[0.5, 1]$ (normalized units).
\end{itemize}

\cref{fig:frequency_analysis} shows this analysis for each of the 4 diffusion models compared to the MS COCO real images.
It is evident how the diffusion process acts similarly to a low-pass filter, attenuating high-frequency components.
It is nevertheless clear how different models have different frequency responses, with SDXLT having the steepest decay of high-frequency components.
As this might be describe as a consequence of the deeply different training procedures and objective of each model (e.g., the strong compression achieved via distillation on SDXLT for low-steps inference), we observe this might also correlate with diversity and distribution shift, measured via FID score, with the only exception of SANA, which has a less steep decay but the lowest recorded diversity.
This analysis suggests that the frequency characteristics of generated images could be a contributing factor to the recognizability-diversity tradeoff observed earlier, and consequently to the downstream performance of models trained on such data.
Further investigation is although needed to fully understand these relationships.

\section{Hard-negative-aware loss functions}

\begin{figure}[t]
    \centering
    \includegraphics[width=0.4\textwidth]{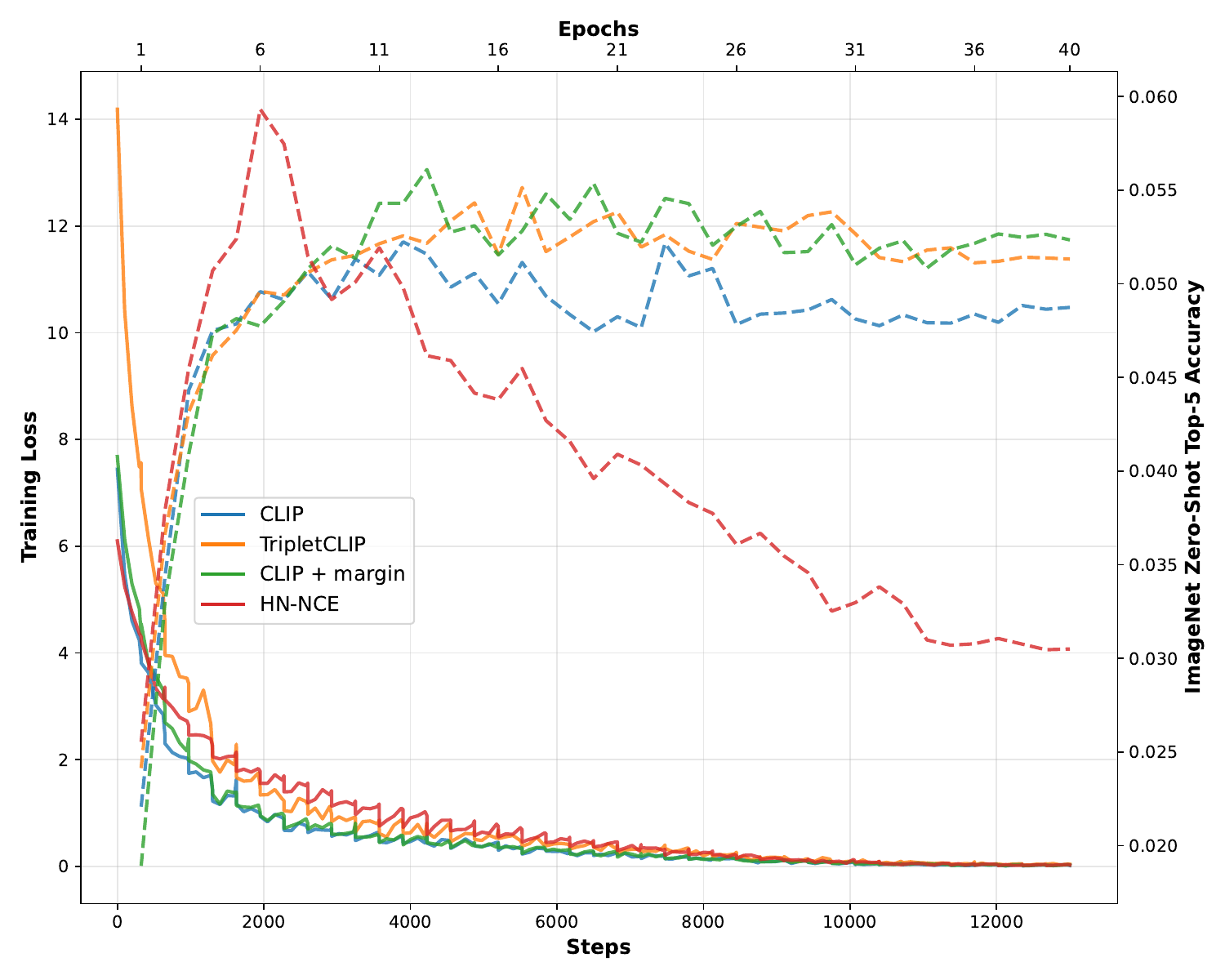}
    \caption{\textbf{Loss function dynamics.} Validation curves show TripletCLIP outperforms standard objectives by explicitly ignoring noisy $T \to I$ comparisons, effectively filtering out false negatives introduced by T2I misalignment.}
    \label{fig:hn_loss_ablation}
\end{figure}

Figure~\ref{fig:hn_loss_ablation} presents an ablation on a set of hard-negative-aware loss funcions comparing $\mathcal{L}_{CLIP}$, with and without $\mathcal{L}_{margin}$, $\mathcal{L}_{HN-NCE}$, and $\mathcal{L}_{TripletCLIP}$. From these results, it seems that TripletCLIP overcomes the excessively noisy signal introduced by hard negative images while still leveraging hard negatives, preserving the image-to-text direction. Differently, the margin loss also seems to help performance, trying to enforce a relative hierarchy of similarity between the anchor, its hard negatives, and other ``easier'', less similar negatives. This loss can, however, be exploited more effectively when there is a 1-to-many relationship between anchors and hard negatives, which is not the case in our current dataset design. A very different behavior is shown by HN-NCE, whose importance-weighting strategy seems peak at very early training stages, but then quickly degrades, possibly due to an excessively punishing weighting scheme in these settings.

\section{Balanced sampling}
\label{app:balanced_sampling}

\begin{figure}[t]
    \centering
    \includegraphics[width=\textwidth]{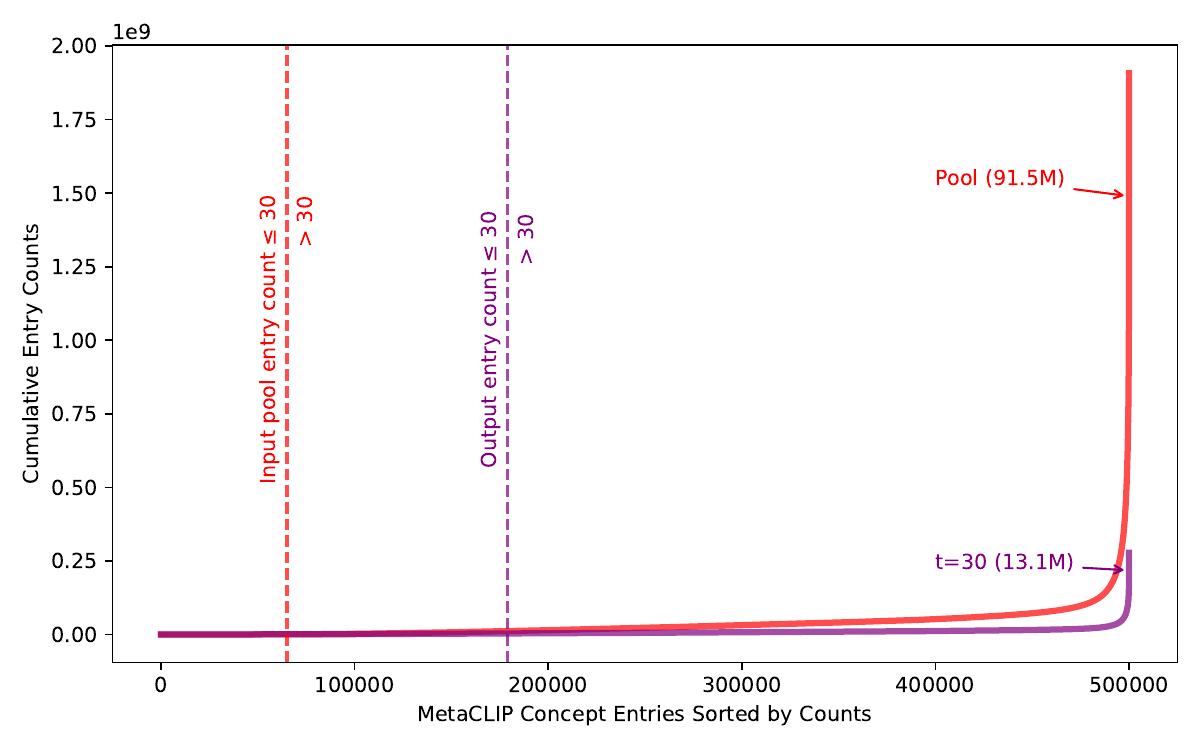}
    \caption{\textbf{Distributional rectification via Balanced Sampling.} The algorithm flattens the power-law distribution typical of web data, increasing the effective sampling rate of tail concepts to ensure uniform semantic coverage.}
    \label{fig:entry_count}
\end{figure}

Balanced sampling is a sampling algorithm proposed by \citea{xu2023demystifying} as a data curation strategy designed to overcome the ``rich-get-richer'' problem found in raw web-crawled datasets, where a small number of common concepts appear millions of times, while rare concepts are underrepresented, leading to a highly skewed distribution that limits a model's ability to generalize. Given a thresholding count $t$, this algorithm assigns a sampling probability to each concept in $C$ equal to $t/count(c)$, where $count(c)$ equals to how many times $c$ appears in the caption set, if $count(c)>t$, and a probability of $1$ otherwise. Each caption is sampled based on the probability of each concept contained in it. We generate an initial caption set of $100M$ captions, then de-duplicate the captions and run this sampling algorithm with $t=30$, leading to a final caption set of $13.1M$ captions. We finally uniformly downsample the caption set to our final target caption set $T^+$ of the desired size. \cref{fig:entry_count} shows how this algorithm manages to apply a strong downsampling to high-frequency concepts.

\section{Delta Multi-Task Learning Metric}
\label{app:deltamtl}
To summarize the overall performance of models capable of performing multiple tasks, aggregating metrics which are not directly comparable, \citea{deltamtl} defined the Delta Multi-Task Learning ($\Delta_{MTL}$) metric, which computes the average relative performance gain or drop of a model $m$ w.r.t. a baseline model $b$ on a set of tasks $T$:
\begin{equation}
    \Delta_{MTL} = 100 \times \frac{1}{T} \sum_{i=1}^{T} (-1)^{l_i} \frac{M_{m,i}-M_{b,i}}{M_{b,i}},
\end{equation}
with $l_i$ being 1 if a lower value means better performance for metric $i$, and 0 otherwise.

\section{Qualitative Examples}
\label{app:qualitative}

\begin{figure}[t]
    \centering
    \includegraphics[width=\textwidth]{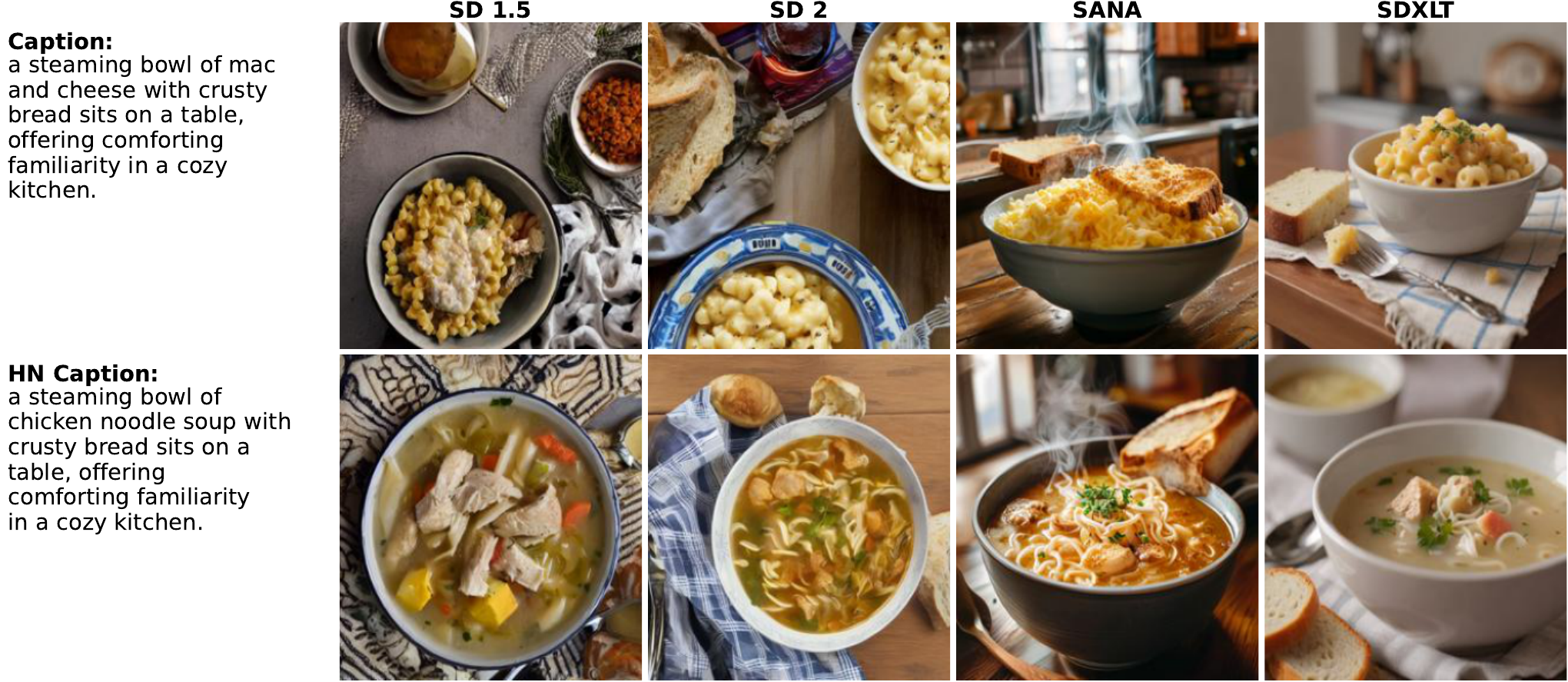}
    \includegraphics[width=\textwidth]{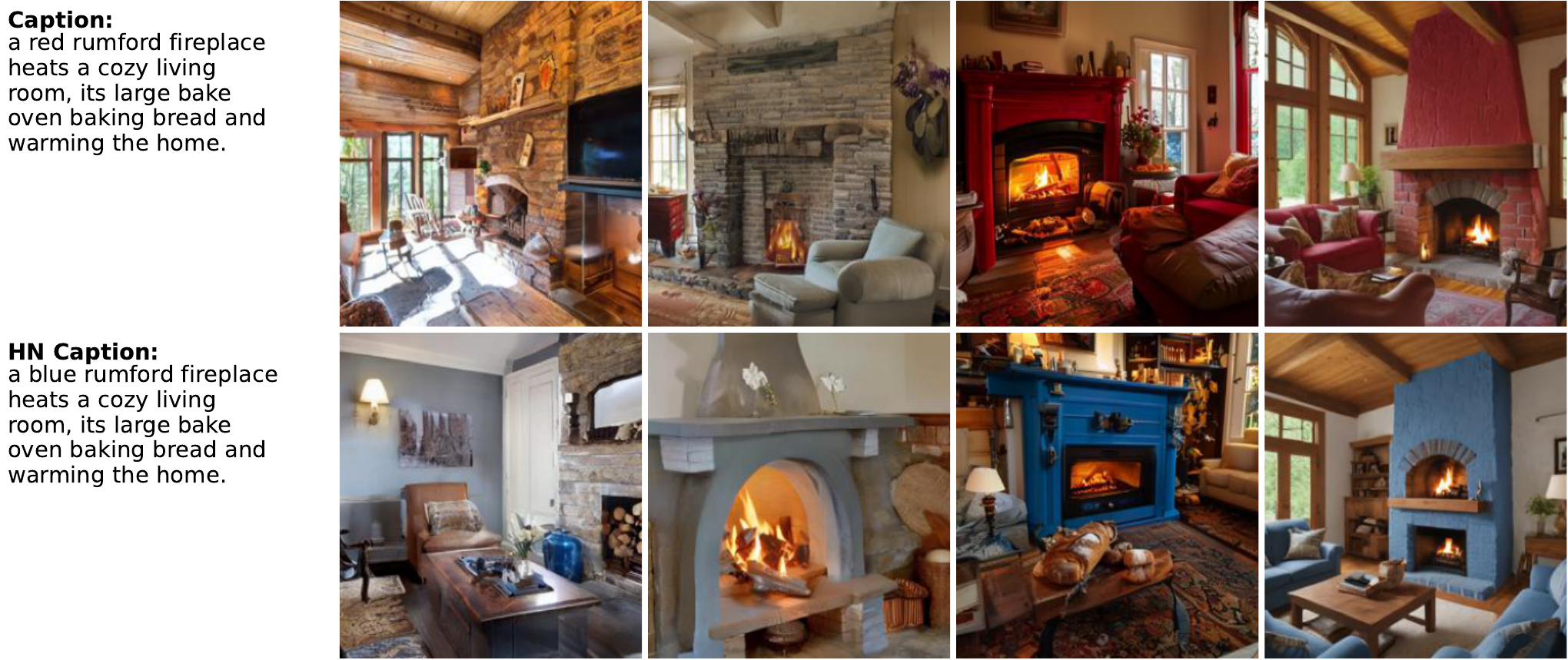}
    \includegraphics[width=\textwidth]{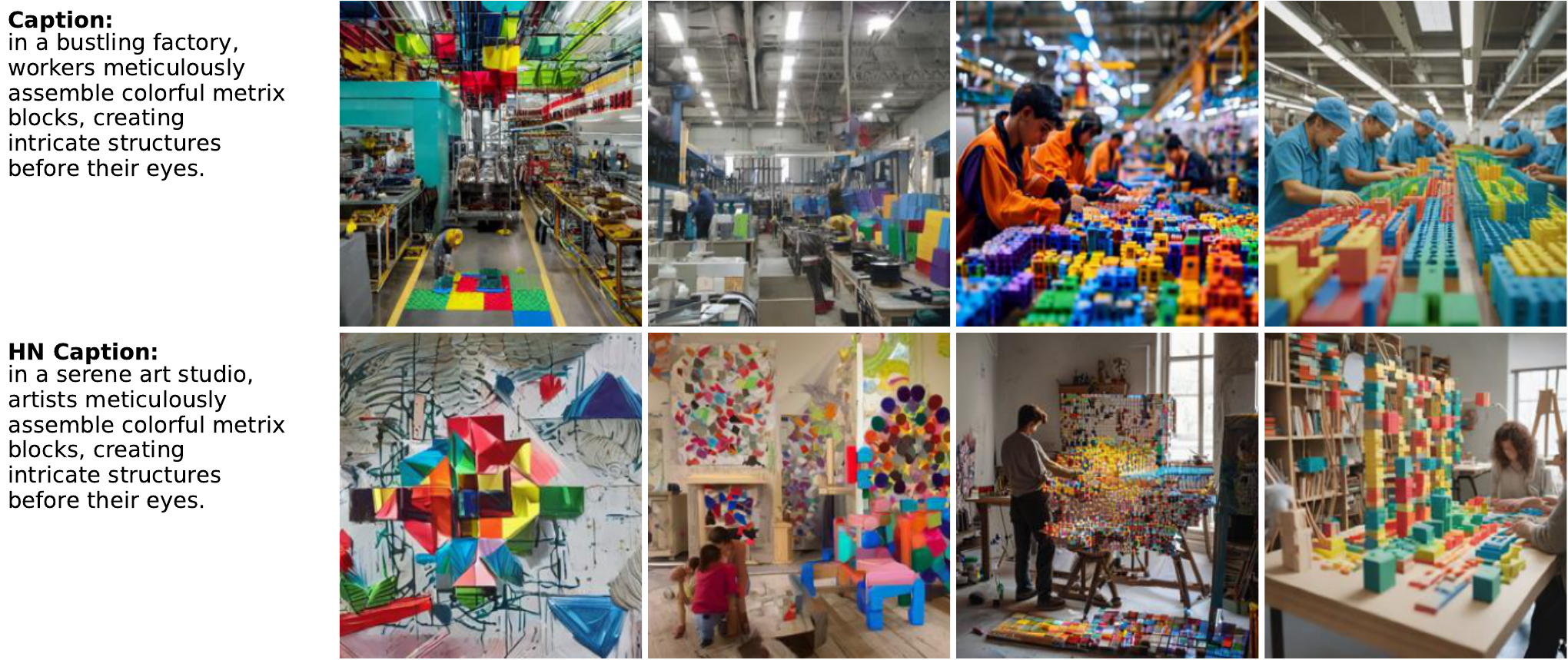}
    \caption{\textbf{Visualizing Generator Invariance.} The ensemble renders identical semantic shifts across disjoint visual manifolds, forcing the encoder to align with the shared concept modification rather than generator-specific stylistic cues.}
    \label{fig:synthmphn_examples}
\end{figure}

\begin{figure}[t]
    \centering
    \includegraphics[width=\textwidth]{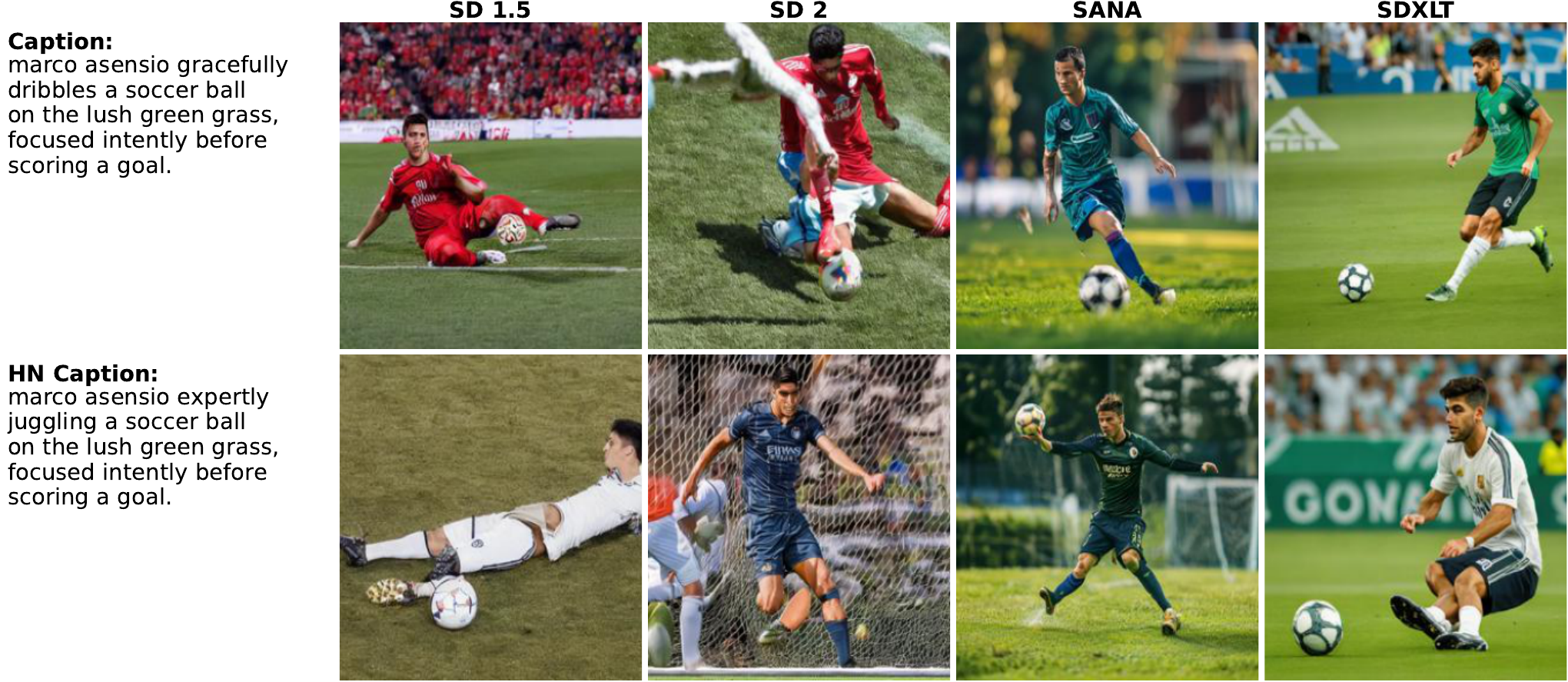}
    \includegraphics[width=\textwidth]{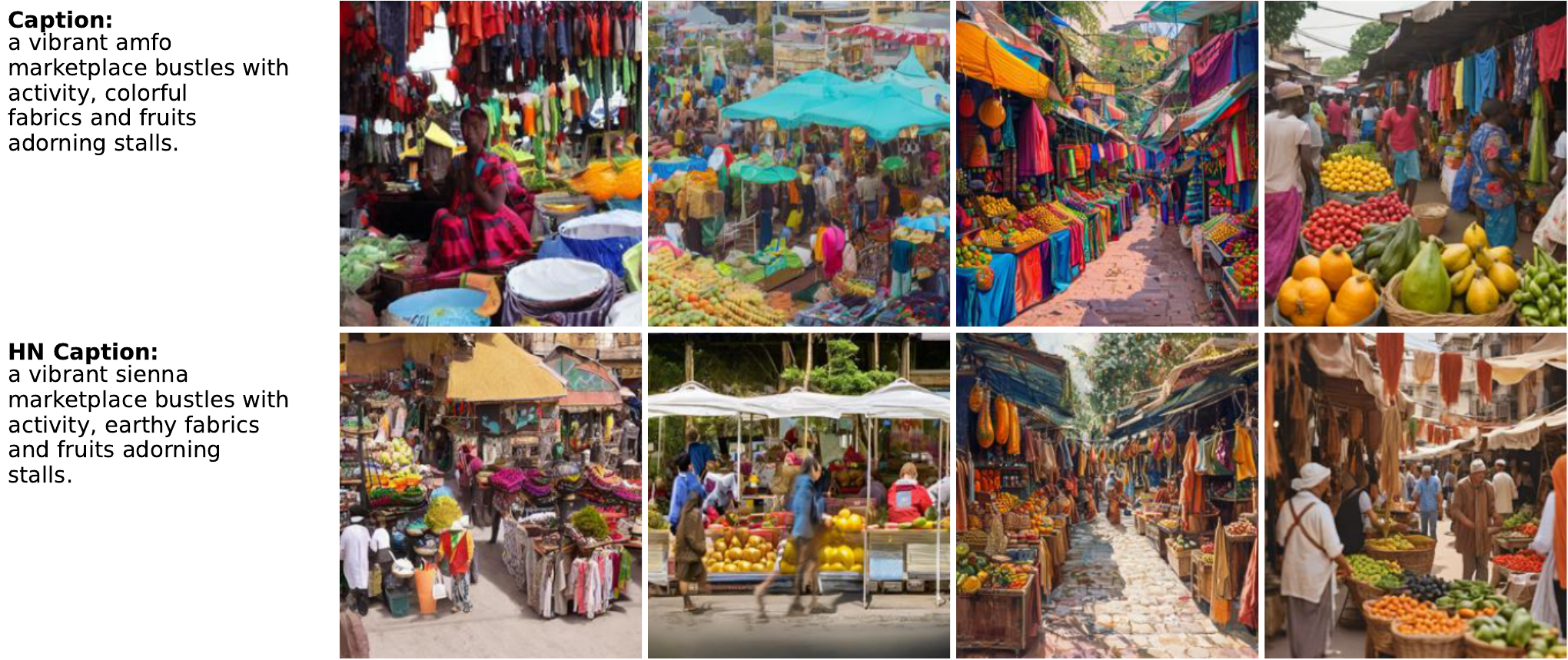}
    \includegraphics[width=\textwidth]{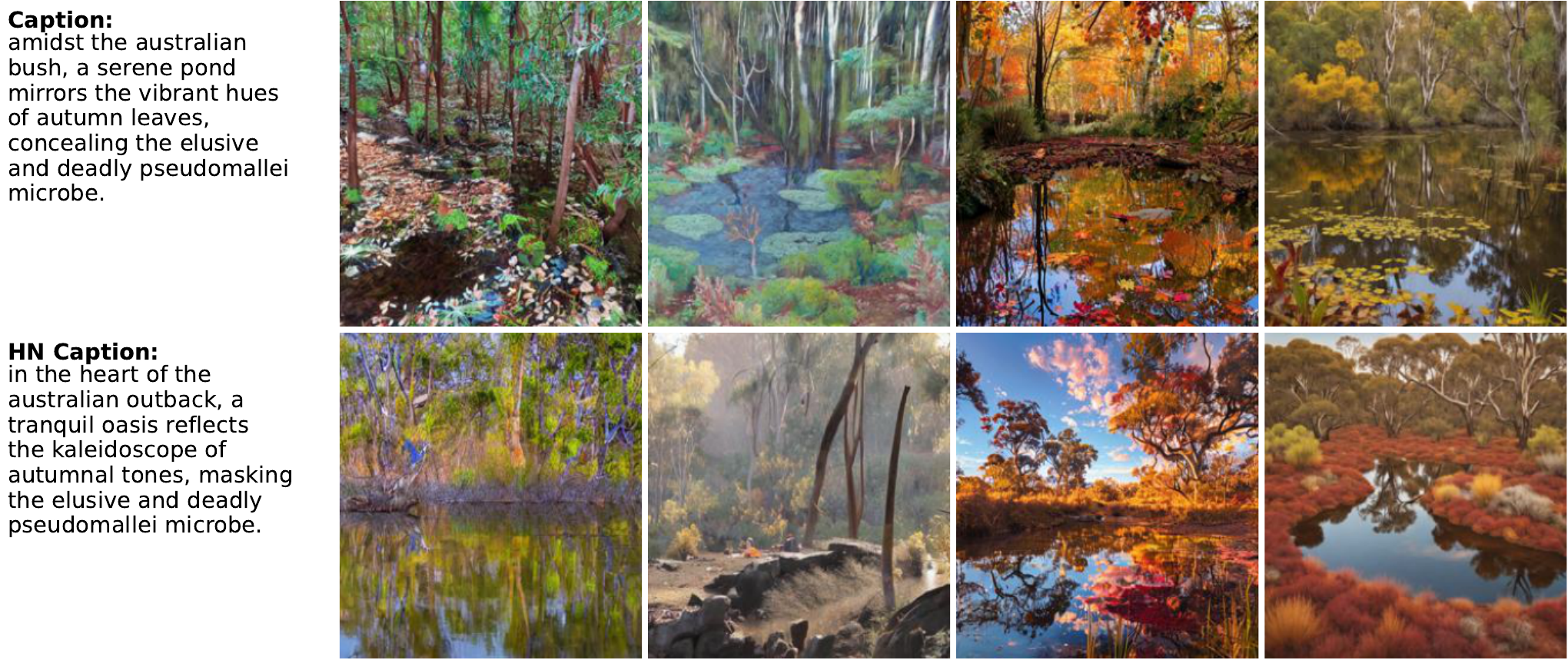}
    \caption{\textbf{Failure modes and Ensemble resilience.} Instances of attribute leakage or visual ambiguity in single generators. The multi-generator approach mitigates these local errors by averaging the learning signal across the ensemble.}
    \label{fig:synthmphn_examples_fail}
\end{figure}

We present qualitative examples to visually assess the effectiveness of our hard negative generation pipeline and the impact of the multi-generator ensemble. \cref{fig:synthmphn_examples} highlights successful generations where the distinct semantic modification is correctly reflected across different architectures. Conversely, \cref{fig:synthmphn_examples_fail} illustrates failure cases where specific generators struggle with fine-grained attribute binding or produce ambiguous visualizations, underscoring the necessity of our ensemble approach to marginalize individual model errors.

\end{document}